\pdfoutput=1
\documentclass{article}

\usepackage{PRIMEarxiv}   % arXiv style file; if your .sty has another name (e.g. arxiv.sty), change this line

\usepackage[utf8]{inputenc}
\usepackage[T1]{fontenc}
\usepackage{amsmath,amssymb,bm}
\usepackage{booktabs,tabularx,threeparttable,array,multirow}
\usepackage{colortbl,xcolor}
\usepackage{setspace}
\usepackage{graphicx}
\usepackage{url}
\usepackage{microtype}
\usepackage{tikz}
\usetikzlibrary{positioning,fit,backgrounds,arrows.meta}
\usepackage{algorithm}
\usepackage{algpseudocode}
\usepackage[round,authoryear]{natbib}
\usepackage[hidelinks]{hyperref}
\hypersetup{
    colorlinks=true,
    breaklinks=true,
    urlcolor=blue,
    linkcolor=red,
    bookmarksopen=false,
    filecolor=black,
    citecolor=blue,
    linkbordercolor=pink,
    pdftitle={Do operators need deep learning to forecast shared-bike demand around metro stations? Evidence from Shanghai},
    pdfauthor={Shuxia Pang, Di Zhu, Wenxiang Li, Qiuyu Liao, Maolin Wang, Tianyi Li},
    pdfkeywords={Dockless bike sharing; Metro-bike integration; Short-term demand forecasting; Fleet rebalancing; Graph convolutional network; Statistical baseline}
}

\title{Do operators need deep learning to forecast shared-bike demand around metro stations?\\ Evidence from Shanghai}

\author{%
  Shuxia Pang\textsuperscript{1}, Di Zhu\textsuperscript{2}, Wenxiang Li\textsuperscript{3}, Qiuyu Liao\textsuperscript{1}, Maolin Wang\textsuperscript{4}, Tianyi Li\textsuperscript{1,*}\\[1.5ex]
  \parbox[t]{0.92\textwidth}{\centering\small
    \textsuperscript{1}Department of Civil Engineering, Saint Louis University, St.~Louis, MO 63103\\
    \textsuperscript{2}Department of Geography, University of Minnesota, Twin Cities, MN 55455, United States\\
    \textsuperscript{3}Business School, University of Shanghai for Science and Technology, Shanghai, 200093, China\\
    \textsuperscript{4}Civil Aviation Management Institute of China, Beijing, 100102, China\\[1ex]
    \textsuperscript{*}Corresponding author: \href{mailto:tianyili.ai@gmail.com}{\texttt{tianyili.ai@gmail.com}}%
  }%
}

\begin{document}
\maketitle

\begin{abstract}
Bike-sharing operators rely on short-term demand forecasts to rebalance fleets around metro stations, and cities rely on them to manage parking at station entrances. Deep learning models that combine metro ridership with bike activity are increasingly proposed for this task, but the evidence behind them rarely separates the value of the metro data from the value of the model and seldom includes a well-tuned statistical baseline. Operators and transit agencies therefore have little basis for deciding which forecasting system to build or whether to share metro data. Using 20 days of metro fare-collection records and 18.06 million dockless-bike records for 342 stations in Shanghai, we compare a transparent statistical pipeline, three deep learning models with and without metro inputs, and combinations, under one evaluation protocol with held-out test days. Metro data reduce forecast error by 1--3\% at station level but by 10--34\% for the network total. The choice among deep learning models changes station-level error by at most 3\%. The statistical pipeline is as accurate as the best deep learning model at station level, has the lowest mean error for the network total and alone provides calibrated prediction intervals; its error is also 16--41\% below the published scores of a graph-based model for the same raw data. For operators, a transparent statistical pipeline is a sufficient default, with deep learning as a complement at the busiest stations. For transit agencies and cities, metro data are valuable mainly for forecasting demand across the whole fleet. Accuracy claims for forecasting tools should be tested against a tuned statistical baseline.
\end{abstract}

\keywords{Dockless bike sharing \and Metro--bike integration \and Short-term demand forecasting \and Fleet rebalancing \and Graph convolutional network \and Statistical baseline}

%=====================================================================
\section{Introduction}
\label{sec:intro}
%=====================================================================

Urban travelers increasingly organize their daily mobility as multimodal chains in which fixed-schedule rail systems deliver them to high-capacity nodes and flexible micromobility services redistribute them across the finer-grained urban fabric \citep{martin2014evaluating,zhao2017bicycle,jin2025connecting}. Shared bikes near metro stations therefore function less as a standalone mode than as the behavioral extension of a rail trip: the first- or last-mile segment through which a commuter completes an origin-to-destination chain \citep{ma2015bicycle,bocker2020bike}, extending the effective catchment of rail stations and reducing reliance on private vehicles or congested bus feeders. The operational value of these systems depends on predicting short-term demand accurately enough to enable proactive fleet rebalancing \citep{raviv2013optimal,faghih2017empirical,dell2018bike,gammelli2022predictive}, and the need is most acute near metro exits, where flows peak sharply with commuting rhythms and train arrivals. City authorities have a parallel interest: the regulation of dockless fleets in China passed within a few years through phases of free growth, regulation and limitation \citep{gu2019dockless}, and parking where bikes concentrate remains a concern in the management of dockless micromobility \citep{zhao2021geo,brousseau2025harmonizing}.

The forecasting literature has responded with increasingly elaborate spatiotemporal deep networks \citep{yin2022deep,jiang2022graph}. Recurrent units capture periodic patterns \citep{ma2015long}, graph convolutions propagate demand along station connections \citep{lin2018predicting,yu2018spatio}, and multi-task or multi-graph designs fuse heterogeneous modal signals \citep{liang2022joint,ke2021predicting}. Metro ridership is closely tied to feeder-mode use near transit hubs \citep{yang2026integrating}, but it is rarely used as an explicit dynamic input. The studies that do use it report accuracy gains \citep{liang2022bike}. A companion study on Shanghai data by part of the present team \citep{li2026cogclstm} combines graph convolution and long short-term memory with metro entry/exit flows and reports reductions in weighted absolute percentage error (WMAPE) of 22--36\% relative to an LSTM and of 24--35\% relative to the historical average across four aggregation intervals and both flow directions. Results of this kind are usually taken to show that metro ridership is a valuable predictor of bike demand and that graph-based architectures are needed to exploit it.

The typical experiment does not support either conclusion on its own. When a new model is given new inputs and compared with generic baselines, three sources of improvement are confounded: the information added (metro flows), the architecture that uses it (graph convolution, multi-task heads) and the evaluation protocol (data split, tuning, metrics). Studies in the general forecasting literature have repeatedly found that, when the protocol is held fixed and the baselines are carefully specified, sophisticated models add little over statistical methods \citep{makridakis2018statistical,makridakis2020m4,elsayed2021really,zeng2023transformers}. Whether the same holds for metro-coupled bike demand, where the cross-modal structure that deep models are designed to exploit is strong, has not been examined. The answer has practical consequences. It determines whether operators should invest in graph-based forecasting infrastructure or in a statistical model that runs in seconds and provides predictive intervals, whether a real-time metro data feed is worth establishing between a transit agency and an operator, and what evidence those who procure or review forecasting tools should ask for.

This paper addresses three questions on a single dataset under a single protocol: how much metro information adds once the architecture is held fixed, and whether the answer depends on the scale at which demand is forecast (station or network); how much the architecture adds once the information is held fixed; and whether the forecasts of the deep models carry information beyond an intraday profile scaled by recent activity, or reproduce that structure at greater cost. The aim is to measure what each model family contributes to this application and under which conditions.

\begin{figure}[htbp]
\centering
\begin{tikzpicture}[
  x=1cm, y=1cm, font=\scriptsize,
  panel/.style={draw=fwLine, fill=fwFill, rounded corners=3pt, line width=0.5pt},
  htxt/.style={text=fwHeadTxt, font=\footnotesize\bfseries, inner sep=0pt, anchor=west},
  stxt/.style={font=\scriptsize\bfseries, text=fwNavy, inner sep=0pt, anchor=west},
  txt/.style={align=left, text width=4.05cm, inner sep=0pt, anchor=north west},
  mini/.style={draw=fwNavy, fill=white, rounded corners=1.5pt, line width=0.4pt, inner xsep=1.5pt, inner ysep=2.5pt, font=\fontsize{6pt}{7pt}\selectfont, align=center},
  arr/.style={-{Stealth[length=5pt,width=4.5pt]}, line width=0.9pt, draw=fwNavy},
  tarr/.style={-{Stealth[length=3pt,width=2.5pt]}, line width=0.4pt, draw=fwNavy},
  ax/.style={-{Stealth[length=2.5pt,width=2pt]}, line width=0.4pt, draw=fwGrey},
  lbl/.style={font=\tiny, text=fwGrey, inner sep=1pt},
  plbl/.style={font=\fontsize{6pt}{7pt}\selectfont, inner sep=1pt, anchor=north, align=center}
]
% ---- colour scheme: light-grey headers, black text, blue/vermilion for data
\definecolor{fwHead}{HTML}{E6E6E6}
\definecolor{fwHeadTxt}{HTML}{1A1A1A}
\definecolor{fwNavy}{HTML}{1A1A1A}
\definecolor{fwBlue}{HTML}{0072B2}
\definecolor{fwLight}{HTML}{9E9E9E}
\definecolor{fwOrange}{HTML}{D55E00}
\definecolor{fwLine}{HTML}{BDBDBD}
\definecolor{fwFill}{HTML}{FFFFFF}
\definecolor{fwGrey}{HTML}{5F5F5F}
\definecolor{fwPale}{HTML}{F0F0F0}

% panel: width 7.85, height 5.0; header 0.55; left column 0.25--3.3, divider 3.5, text column 3.65--7.7
\newcommand{\panelbox}[3]{%
  \begin{scope}[shift={(#1,#2)}]
    \path[panel] (0,0) rectangle (7.85,5.0);
    \path[fill=fwHead, rounded corners=3pt] (0,4.45) rectangle (7.85,5.0);
    \path[fill=fwHead] (0,4.45) rectangle (7.85,4.7);
    \node[htxt] at (0.3,4.725) {#3};
    \draw[fwLine, line width=0.4pt] (3.5,0.35) -- (3.5,4.1);
  \end{scope}}

% ================================================================ 1 data
\panelbox{0}{5.5}{1\quad Data}
\begin{scope}[shift={(0,5.5)}]
  % station and catchments (left)
  \draw[fwLight, dashed, line width=0.4pt] (1.05,2.35) circle (0.78);
  \fill[fwPale] (1.05,2.35) circle (0.33);
  \draw[fwNavy, dashed, line width=0.5pt] (1.05,2.35) circle (0.33);
  \node[fill=fwNavy, text=white, font=\fontsize{6pt}{6pt}\selectfont\bfseries, inner sep=1.2pt, minimum size=0.22cm] at (1.05,2.35) {M};
  \foreach \p in {(0.85,2.45),(1.25,2.20),(0.95,2.12),(1.28,2.52),(0.78,2.28),(1.15,2.62),(0.60,2.75),(1.55,1.95),(1.45,2.80),(0.55,1.95),(1.62,2.55),(0.72,2.95),(1.30,1.75),(0.42,2.40),(1.67,2.10),(0.80,1.72)}
    \fill[fwBlue] \p circle (0.03);
  \node[lbl, text=fwNavy] at (1.05,1.93) {200\,m};
  \node[lbl] at (1.05,1.45) {1\,km};
  \node[lbl, anchor=north, text width=1.7cm, align=center] at (1.05,1.2) {catchments around each metro station};
  % intraday profile (right)
  \draw[ax] (2.1,1.65) -- (3.35,1.65); \draw[ax] (2.1,1.65) -- (2.1,3.2);
  \node[lbl, anchor=north] at (2.1,1.63) {06\,h}; \node[lbl, anchor=north] at (3.25,1.63) {23\,h};
  \draw[fwOrange, line width=0.8pt] plot[smooth, tension=0.7] coordinates
    {(2.12,1.75) (2.27,1.95) (2.40,2.98) (2.54,2.30) (2.70,2.12) (2.86,2.30) (3.02,2.90) (3.17,2.05) (3.27,1.80)};
  \draw[fwBlue, line width=0.8pt] plot[smooth, tension=0.7] coordinates
    {(2.12,1.70) (2.27,1.82) (2.40,2.52) (2.54,2.05) (2.70,1.95) (2.86,2.05) (3.02,2.48) (3.17,1.90) (3.27,1.72)};
  \draw[fwOrange, line width=0.8pt] (1.95,1.25) -- (2.15,1.25); \node[lbl, anchor=west] at (2.17,1.25) {metro\,AFC};
  \draw[fwBlue, line width=0.8pt] (1.95,1.05) -- (2.15,1.05); \node[lbl, anchor=west] at (2.17,1.05) {bike events};
  % text
  \node[txt] at (3.65,4.15) {\textbf{Shanghai, 12--31 Dec 2022}: 342 metro stations; 5-min AFC entries and exits; 18.06\,M dockless-bike records.\\[3pt]
    \textbf{Main panel}: 200\,m, hourly, 20 days.\\[1pt]
    \textbf{1-km panel}: nearest station, 15/30/60\,min, 336 stations.\\[3pt]
    Demand falls by two thirds; 41\% of station-hours are zero.};
\end{scope}

% ================================================================ 2 coupling
\panelbox{8.2}{5.5}{2\quad Coupling diagnostics}
\begin{scope}[shift={(8.2,5.5)}]
  \fill[fwPale] (0.8,1.25) rectangle (1.76,3.5);
  \draw[ax] (0.8,1.25) -- (3.35,1.25); \draw[ax] (0.8,1.25) -- (0.8,3.7);
  \foreach \x/\l in {0.8/0,1.76/200,3.2/500}
    {\draw[fwGrey, line width=0.3pt] (\x,1.25) -- (\x,1.2); \node[lbl, anchor=north, inner sep=0.5pt] at (\x,1.2) {\l};}
  \node[lbl, anchor=north] at (2.0,0.95) {distance from station (m)};
  \node[lbl, rotate=90, anchor=south] at (0.6,2.35) {elasticity $\gamma_r$};
  \draw[fwLight, line width=0.4pt, dashed] (0.8,2.1) -- (3.3,2.1);
  \node[lbl, anchor=north west] at (0.86,2.07) {half of peak};
  \draw[fwNavy, line width=0.5pt] (1.04,2.85) -- (1.40,2.95) -- (1.64,2.80) -- (2.00,2.45) -- (2.48,2.03) -- (2.96,1.82);
  \foreach \x/\y/\e in {1.04/2.85/0.12,1.40/2.95/0.10,1.64/2.80/0.13,2.00/2.45/0.12,2.48/2.03/0.12,2.96/1.82/0.10}
    {\draw[fwNavy, line width=0.4pt] (\x,\y-\e) -- (\x,\y+\e); \fill[fwNavy] (\x,\y) circle (0.045);}
  \node[lbl, text=fwNavy, anchor=south west] at (0.86,3.2) {flat to 200\,m};
  \node[lbl, anchor=north east] at (3.3,0.68) {(schematic)};
  \node[txt] at (3.65,4.15) {\textbf{PPML elasticities} on rings and disks, 0--500\,m; station, hour-of-day and date fixed effects; 60/15/5\,min.\\[3pt]
    \textbf{Granger causality}, hourly, by station.\\[3pt]
    $\Rightarrow$ Coupling is short-range: flat to 200\,m, half of peak by 300--400\,m; entry$\to$arrival exceeds exit$\to$departure.};
\end{scope}

% ================================================================ 3 models
\panelbox{0}{0}{3\quad Models}
\begin{scope}
  \node[stxt] at (0.3,3.9) {Statistical pipeline};
  \foreach \x/\l in {0.64/a, 1.42/b, 2.20/c, 2.98/d}
    \node[mini, fill=fwNavy, text=white, font=\fontsize{6pt}{6pt}\selectfont\bfseries, minimum size=0.3cm, inner sep=1pt] (s\l) at (\x,3.45) {\l};
  \node[plbl] at (0.64,3.25) {profile};
  \node[plbl] at (1.42,3.25) {EB\\shrink.};
  \node[plbl] at (2.20,3.25) {level\\ratio};
  \node[plbl] at (2.98,3.25) {NB\\errors};
  \draw[tarr] (sa) -- (sb); \draw[tarr] (sb) -- (sc); \draw[tarr] (sc) -- (sd);
  \node[stxt] at (0.3,2.45) {Deep models};
  \draw[fwLight, line width=0.5pt] (0.35,2.12) -- (0.64,1.89) -- (0.91,2.15) -- (1.12,1.87);
  \draw[fwLight, line width=0.5pt] (0.35,2.12) -- (0.57,1.59) -- (0.64,1.89);
  \draw[fwLight, line width=0.5pt] (0.57,1.59) -- (0.91,1.67) -- (1.12,1.87);
  \foreach \p in {(0.35,2.12),(0.64,1.89),(0.91,2.15),(0.57,1.59),(1.12,1.87),(0.91,1.67)} \fill[fwNavy] \p circle (0.055);
  \node[lbl, text=black, anchor=north west, text width=2.05cm, align=left] at (1.26,2.25) {LSTM, ST-GCN,\\MBC-GCN\\\mbox{AR: bike only}\\\mbox{Semi-AR: +metro}};
  \node[stxt] at (0.3,1.0) {Hybrids};
  \node[lbl, text=black, anchor=north west, text width=2.95cm, align=left] at (0.3,0.83) {statistical offset + LSTM residual (SC-LSTM); equal-weight average};
  \node[txt] at (3.65,4.15) {\textbf{Statistical pipeline}: profile $\to$ EB shrinkage $\to$ level ratio $\to$ NB errors; no trainable weights.\\[3pt]
    \textbf{Deep models}: LSTM, ST-GCN, MBC-GCN; AR (bike only) or Semi-AR (bike + metro). Network total: also \mbox{SARIMAX}, ridge, boosting.\\[3pt]
    \textbf{Hybrids}: offset + LSTM residual; equal-weight mean.};
\end{scope}

% ================================================================ 4 evaluation
\panelbox{8.2}{0}{4\quad Evaluation protocol}
\begin{scope}[shift={(8.2,0)}]
  \node[stxt] at (0.3,3.9) {Rolling origin};
  \foreach \f/\y/\t/\v in {1/3.2/12/14, 2/2.9/14/16, 3/2.6/16/18}{%
    \fill[fwLine] (0.55,\y-0.09) rectangle (0.55+0.135*\t,\y+0.09);
    \fill[fwLight] (0.55+0.135*\t,\y-0.09) rectangle (0.55+0.135*\v,\y+0.09);
    \fill[fwNavy] (0.55+0.135*\v,\y-0.09) rectangle (0.55+0.135*\v+0.27,\y+0.09);
    \node[lbl, anchor=east] at (0.5,\y) {\f};}
  \draw[ax] (0.55,2.35) -- (3.3,2.35);
  \node[lbl, anchor=north west] at (0.5,2.33) {12 Dec}; \node[lbl, anchor=north east] at (3.3,2.33) {31 Dec};
  \fill[fwLine] (0.30,1.75) rectangle (0.45,1.87); \node[lbl, anchor=west] at (0.46,1.81) {train};
  \fill[fwLight] (1.28,1.75) rectangle (1.43,1.87); \node[lbl, anchor=west] at (1.44,1.81) {validation};
  \fill[fwNavy] (2.76,1.75) rectangle (2.91,1.87); \node[lbl, anchor=west] at (2.92,1.81) {test};
  \node[lbl, anchor=north west, text width=2.95cm, align=left] at (0.3,1.45) {each fold $\times$ 3 seeds; tuning and early stopping on validation days only};
  \node[txt] at (3.65,4.15) {\textbf{Two scales}: 342 stations jointly and the network total.\\[3pt]
    \textbf{Metrics}: MAE, RMSE, $R^2$, bias ratio, NB log score, 90\% coverage, Poisson reference, WMAPE; paired Wilcoxon tests.\\[3pt]
    \textbf{External check}: 1-km, 17/4-day protocol of \citet{li2026cogclstm}.};
\end{scope}

% arrows between panels
\draw[arr] (7.9,8.0) -- (8.15,8.0);
\draw[arr] (12.125,5.47) -- (12.125,5.25) -- (3.925,5.25) -- (3.925,5.03);
\draw[arr] (7.9,2.5) -- (8.15,2.5);

% ================================================================ findings strip
\path[panel, fill=white, draw=fwNavy, line width=0.6pt] (0,-2.6) rectangle (16.05,-0.5);
\path[fill=fwHead, rounded corners=3pt] (0,-2.6) rectangle (0.6,-0.5);
\path[fill=fwHead] (0.3,-2.6) rectangle (0.6,-0.5);
\node[text=fwHeadTxt, font=\footnotesize\bfseries, inner sep=0pt, rotate=90] at (0.3,-1.55) {Findings};
\foreach \x/\n/\t in {
  0.8/{1--3\% vs.\ 10--34\%}/{MAE reduction from metro inputs: station level vs.\ network total},
  4.55/{$\le 3\%$}/{spread between the three deep backbones; the pipeline matches them, leads for the network total},
  8.3/{0.85}/{pipeline--LSTM residual correlation; a network given the pipeline offset adds nothing},
  11.95/{16--41\% lower WMAPE}/{than the published \mbox{CoGCLSTM} under its own 1-km protocol}}
  {\node[font=\footnotesize\bfseries\boldmath, text=fwNavy, anchor=north west, inner sep=0pt] at (\x,-0.72) {\n};
   \node[txt, text width=3.5cm] at (\x,-1.12) {\t};}
\end{tikzpicture}
\caption{Research framework. Metro AFC and dockless-bike records are assembled into station--time panels (1); coupling diagnostics establish that metro flows are a short-range predictor of bike demand (2); a statistical pipeline, three deep architectures and their hybrids (3) are compared under one rolling-origin protocol (4); the strip summarizes the findings of Sections~\ref{sec:experiments} and \ref{sec:discussion}.}
\label{fig:research_framework}
\end{figure}
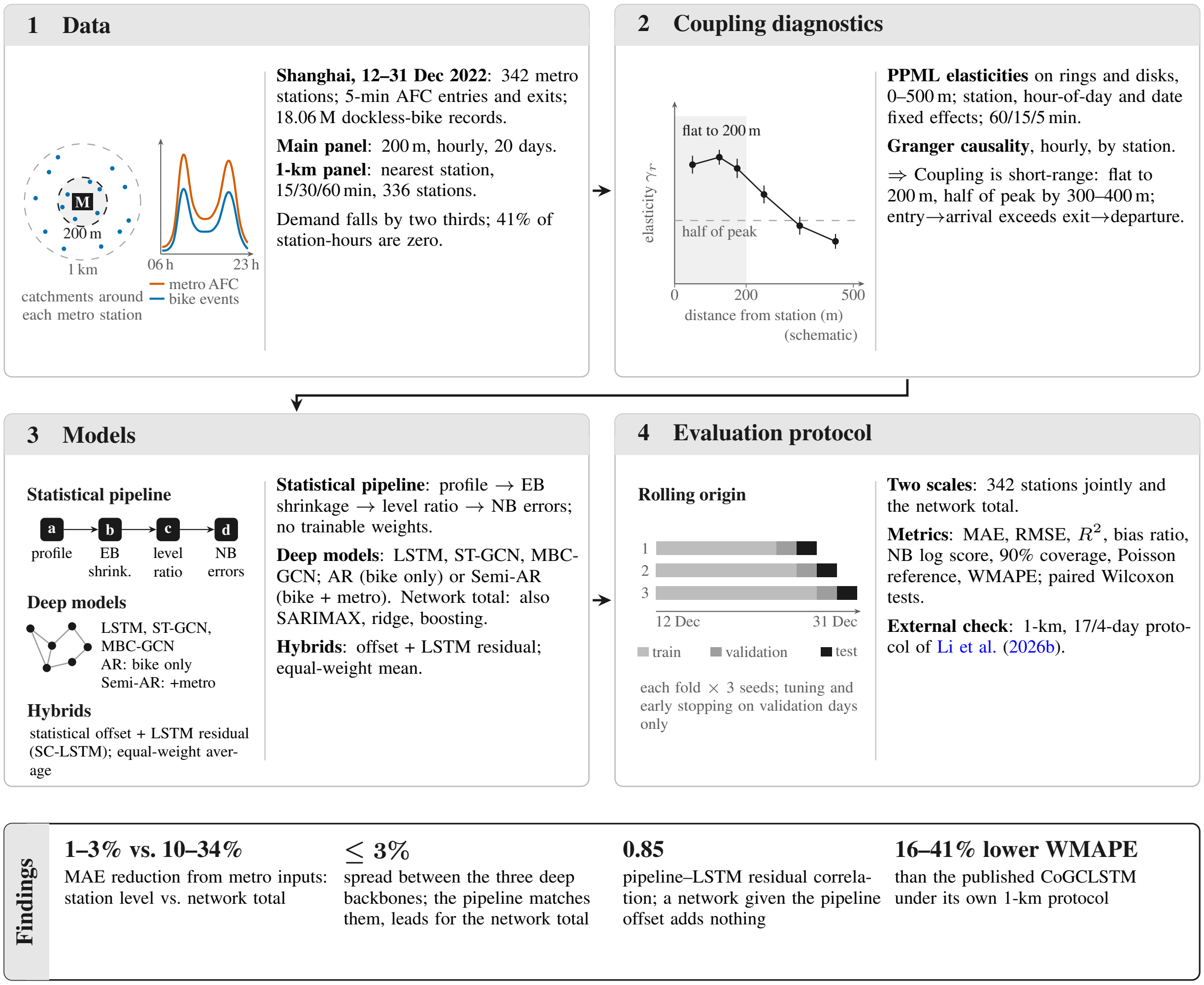

The empirical setting is Shanghai in December 2022, during the COVID-19 wave that followed the relaxation of control measures, with 20 days of synchronized automatic fare collection (AFC) records and dockless-bike lock/unlock events for 342 metro stations. We construct a station--hour panel, document short-run coupling between metro and bike flows, and then run three families of forecasting models under a rolling-origin protocol with held-out test days: a statistical pipeline built from causal recency-weighted profiles, empirical-Bayes shrinkage, a recent-level ratio and a negative-binomial error model; three deep architectures (a long short-term memory network, LSTM; the spatiotemporal graph convolutional network ST-GCN; and a cross-modal graph network, MBC-GCN, short for Metro--Bike Coupling Graph Convolutional Network) crossed with bike-only and bike-plus-metro inputs; and hybrids in which a network corrects the statistical forecast or the two are averaged. Finally, the statistical pipeline is evaluated under the 1-km, 17/4-day protocol of \citet{li2026cogclstm,li2026spatial} on the same raw records, so that the published deep model can be compared against a competent baseline. Figure~\ref{fig:research_framework} summarizes the four stages of the study and its main findings.

The contributions of the paper are as follows. First, the value of metro information for bike-demand forecasting is quantified separately from the architecture and shown to depend on scale: it is small but statistically significant at station level and large for the network total. Second, on a city-scale panel of this length, the choice between a graph and a recurrent backbone is shown to change station-level error by at most a few percent, less than the gap between any learning model and the naive baselines and comparable to the run-to-run variability across folds and seeds. Third, a calibrated statistical pipeline is shown to be as accurate as the most accurate deep model at station level, to have the lowest mean error for the network total and, unlike the deep models, to provide predictive intervals; a network given the statistical forecast as an offset finds almost no residual to learn, and on a panel rebuilt under a published protocol the weighted absolute percentage error of the pipeline is 16--41\% below the published scores of the graph model. For practice, these results indicate that the marginal value of deep learning on this problem lies in combination with a statistical core, and that this value is largest at high-volume stations. The implications for operators, for transit agencies and cities, and for the evaluation of forecasting tools are set out in Section~\ref{sec:implications}.

The remainder of the paper is organized as follows. Section~\ref{sec:lit} positions the study in the literature. Section~\ref{sec:data} describes the data, the coupling diagnostics that motivate the use of metro flows, and the evaluation protocol. Section~\ref{sec:method} specifies the statistical pipeline, the deep models and the hybrids. Section~\ref{sec:experiments} reports the results, Section~\ref{sec:discussion} interprets them and draws the implications for policy and practice, and Section~\ref{sec:conclusion} concludes.

%=====================================================================
\section{Literature review}
\label{sec:lit}
%=====================================================================

Research on metro--bike integration has expanded along two largely parallel tracks: behavioral analyses of first--last mile travel, and algorithmic advances in demand forecasting. A third, smaller literature on forecast evaluation bears directly on how the two tracks should be compared, and a fourth concerns the operational and regulatory decisions that forecasts are meant to support.

\subsection{Metro--bike integration as a dynamic coupled system}
\label{sec:lit_coupling}

Metro systems function as the high-capacity backbone of urban commuting, concentrating passenger demand at fixed spatial nodes, while shared bicycles operate as flexible feeder modes that redistribute demand across finer spatial scales. Empirical studies in high-density cities consistently find that bike-sharing activity around rail stations follows commuting rhythms, station-area built environment and distance to station exits \citep{zhao2017bicycle,guo2020built,guo2021role}, and bike-sharing ridership is higher on routes that connect to metro or rail \citep{bocker2020bike}. Peak-hour synchronization is pronounced: bike arrivals intensify near stations during the morning commute and departures surge as passengers disperse in the evening. Most of this evidence is contemporaneous or daily, estimated through Poisson or negative-binomial regressions, geographically weighted regression or structural equation models, and implicitly assumes a one-directional metro$\to$bike influence. The lag structure of the interaction at the five-minute scale, the reverse influence of bike availability on metro access, and the access/egress asymmetry embedded in commuting rhythms are less well characterized. Data integration is part of the reason: AFC systems record entries and exits at fixed intervals and fixed locations \citep{pelletier2011smart,ma2013mining}, whereas dockless platforms log asynchronous lock/unlock events at continuous coordinates \citep{li2023improving}, and the two are usually analyzed separately or at coarse aggregation. Even deep forecasting models of bike demand \citep{lin2018predicting,li2023improving} rely on historical bike demand and do not use metro ridership as an input.

\subsection{Spatiotemporal demand prediction methods}
\label{sec:lit_forecast}

Short-term demand forecasting has progressed from statistical baselines (historical averages, ARIMA, tree ensembles) to recurrent networks that capture periodicity \citep{ma2015long,xu2018station} and to graph-based models that represent the transport network as a non-Euclidean domain \citep{bronstein2017geometric,shuman2013emerging}. Graph convolution \citep{defferrard2016convolutional,kipf2017semi} combined with temporal convolution or recurrence underlies ST-GCN \citep{yu2018spatio}, DCRNN \citep{li2018diffusion} and many successors \citep{yin2022deep,jiang2022graph}, and has been applied to station-level bike demand \citep{lin2018predicting} and to ride-sourcing and multimodal demand \citep{ke2021predicting,liang2022joint}. Metro ridership enters these models rarely, and when it does it is usually a static proximity feature rather than a dynamic exogenous input \citep{xing2025demand}. The closest antecedent to the present study is \citet{li2026cogclstm}, who fuse metro entry/exit flows, bike events, built-environment indicators and weather in a graph-convolution--LSTM model on the Shanghai data used here and report substantial gains over seven baselines. In that study the baselines do not receive metro inputs, the data are split once into 17 training and 4 test days (a validation set is mentioned for the hyperparameter search but not specified), and the historical-average baseline improves little as the aggregation window lengthens from 15 to 60 minutes. These three features make the size of the architectural contribution difficult to isolate and motivate the fixed-protocol design adopted here.

\subsection{Statistical baselines and forecast evaluation}
\label{sec:lit_eval}

A parallel literature in forecasting science has examined under which conditions complex models outperform simpler ones. The M4 competition found that most pure machine-learning entries were outperformed by well-tuned statistical methods and that gains came mostly from combinations \citep{makridakis2020m4,makridakis2018statistical}; \citet{elsayed2021really} showed that a gradient-boosted model with simple features matched or beat deep architectures on standard benchmarks; \citet{zeng2023transformers} showed that a one-layer linear model outperformed transformer-based forecasters on long-horizon tasks. These findings do not transfer automatically to spatiotemporal count data with strong cross-modal structure, but they motivate the design of the present study: fix the protocol, build a strong statistical baseline, and attribute improvements to information, architecture and combination separately. The evaluation protocol also matters. Rolling-origin evaluation with a held-out test set \citep{tashman2000out,bergmeir2012use}, scale-free error measures for comparison across datasets \citep{hyndman2006another} and paired significance tests \citep{hewamalage2023forecast} are standard recommendations that are not yet standard in the cross-modal forecasting literature; a noise reference against which errors can be judged is rarer still. Finally, forecast combination is one of the most robust results in the field \citep{bates1969combination}. Averaging forecasts with partly independent errors improves accuracy even when neither dominates, which provides a simple diagnostic of whether two models have learned the same thing.

\subsection{Forecasts in operations and policy}
\label{sec:lit_practice}

Forecasts matter to practice through the decisions they feed. For operators the decision is rebalancing: usage and the amount of rebalancing that a system requires vary with station context \citep{faghih2017empirical}, forecast accuracy and the quality of the resulting inventory decisions are distinct \citep{gammelli2022predictive}, and simulation evidence indicates that the value of precise demand forecasts differs between systems \citep{atacc2025bike}. For public authorities the concerns are the regulation of dockless fleets, which in China passed through phases of free growth, regulation and limitation \citep{gu2019dockless}, and the management of parking, for which geo-fenced spaces have been proposed \citep{zhao2021geo} and which remains a barrier to the acceptance of dockless micromobility \citep{brousseau2025harmonizing}. Which forecasting approach and which data a practitioner should adopt, and on what evidence, has received less attention, and this is the question addressed here.

%=====================================================================
\section{Data, coupling diagnostics and evaluation protocol}
\label{sec:data}
%=====================================================================

This section introduces the integrated dataset, the analyses establishing that metro and bike flows are coupled at short lags and over what distance, and the protocol under which all forecasting models are compared.

\subsection{Data preparation and integration}
\label{sec:data_prep}

\begin{figure}[htbp]
  \centering
  \includegraphics[width=0.8\columnwidth]{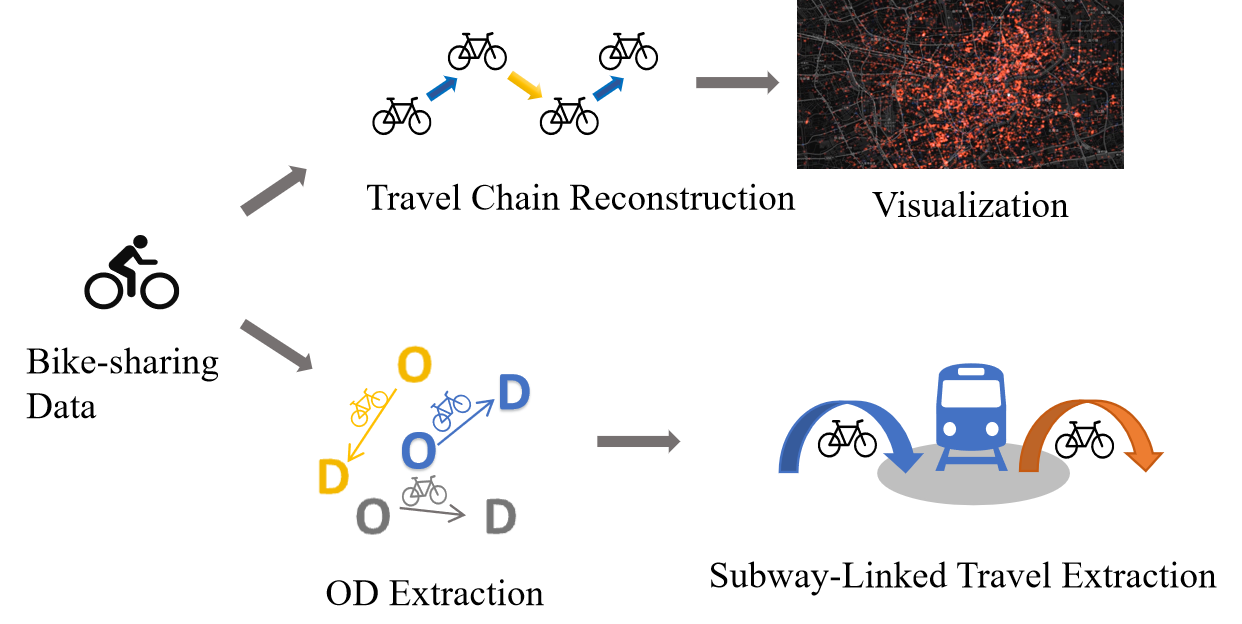}
  \caption{Processing flow of shared-bike records for subway-linked travel extraction.}
  \label{fig:bike_chain}
\end{figure}

\begin{figure}[htbp]
  \centering
  \includegraphics[width=1\columnwidth]{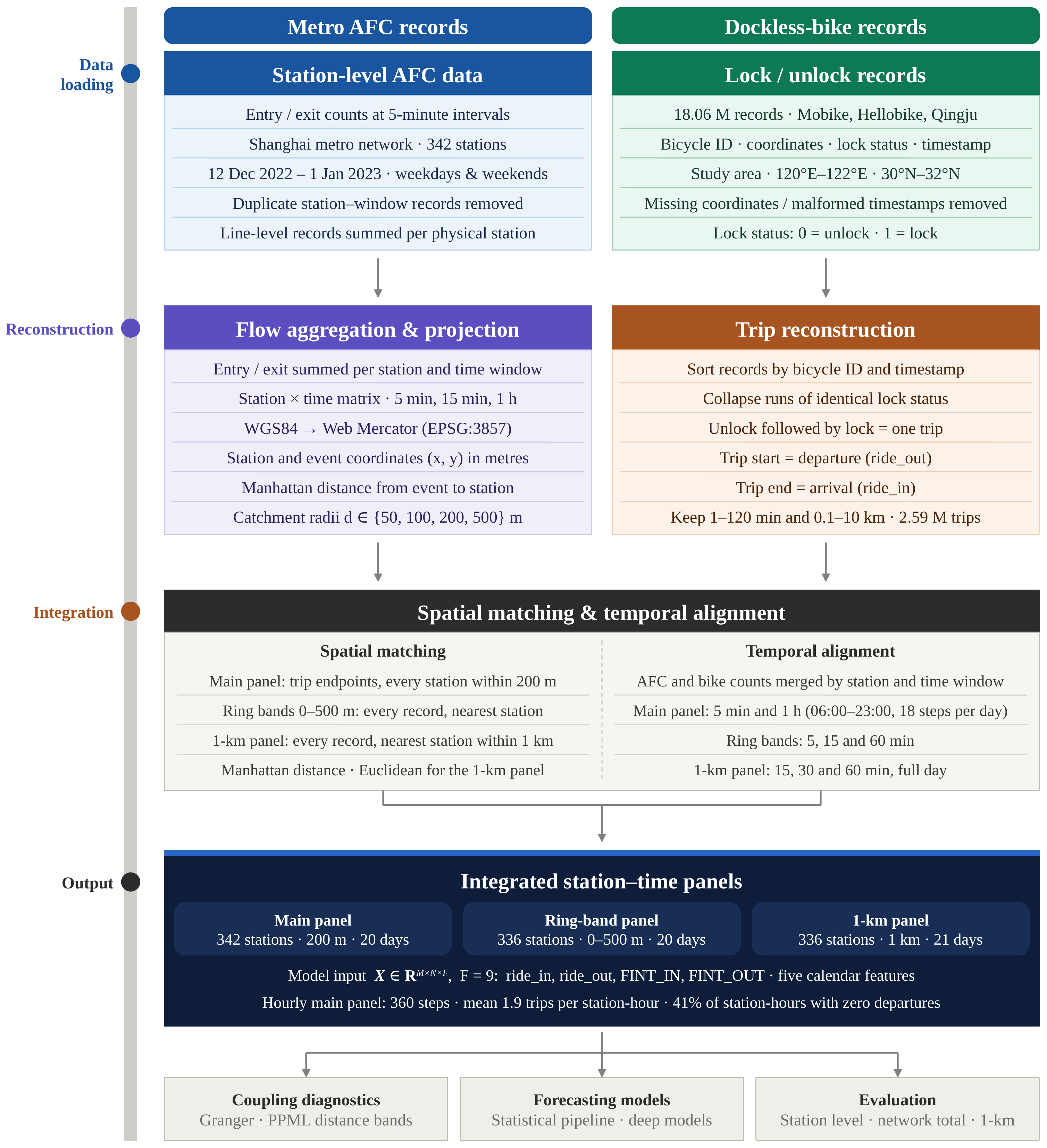}
  \caption{Data integration pipeline from raw metro AFC and bike-sharing records to the integrated station--time panel.}
  \label{fig:data_pipeline}
\end{figure}

We integrate metro AFC passenger flow data with shared-bike operation logs for Shanghai over 12 December 2022 to 1 January 2023, covering both weekdays and weekends within a winter observation window; the main panel uses the 20 days 12--31 December and the 1-km panel of Section~\ref{sec:external} all 21 days. The period coincides with the COVID-19 infection wave that followed the relaxation of control measures in early December 2022. Metro AFC data provide station-level entry and exit counts (\texttt{FINT\_IN}, \texttt{FINT\_OUT}) at 5-minute intervals for the complete network. Dockless-bike data comprise 18.06 million lock/unlock records from the three largest operators (Mobike 60.5\%, Hellobike 30.7\%, Qingju 8.8\%), each carrying a bicycle identifier, coordinates, lock status (0 = unlock, 1 = lock) and a timestamp. Records were restricted to the Shanghai metropolitan area (120$^\circ$E--122$^\circ$E, 30$^\circ$N--32$^\circ$N); records with missing coordinates or malformed timestamps were removed. For each bicycle, records were sorted chronologically and runs of identical lock status collapsed to their first record, so that an unlock followed by a lock on the same bicycle defines one trip: its start is a departure (\texttt{ride\_out}) and its end an arrival (\texttt{ride\_in}). Trips shorter than 1~min or longer than 120~min, or shorter than 100~m or longer than 10~km, were discarded as GPS jitter or non-trips, leaving 2.59 million trips (median duration 9.8~min). Figure~\ref{fig:bike_chain} illustrates the processing of raw bike records and Figure~\ref{fig:data_pipeline} summarizes the integration pipeline.

Events were attributed to metro stations by distance and aggregated to station--time counts, yielding two panels. The \emph{main panel} covers $N=342$ stations at 5-minute and hourly resolution with four flow variables (\texttt{ride\_in}, \texttt{ride\_out}, \texttt{FINT\_IN}, \texttt{FINT\_OUT}) and uses a nominal 200~m catchment, an event being counted at every station within 200~m of it: the distance-band regressions of Section~\ref{sec:coupling_results} show that the elasticity of bike flow with respect to metro flow is flat from the station out to 200~m and halves by 300--400~m, so 200~m is the largest radius that lies wholly inside the coupling plateau; at the hourly resolution used for forecasting it is also the disk with the highest elasticity, and it retains about three times the events of a 100~m disk.\footnote{Distances in the main panel and in the coupling diagnostics are Manhattan distances in Web-Mercator meters. At Shanghai's latitude Web-Mercator distances exceed ground distances by a factor of $1/\cos(31.2^\circ)\approx1.17$, so the nominal 200~m corresponds to about 170~m on the ground; all radii in Section~\ref{sec:coupling_results} are nominal in the same sense. Model comparisons are unaffected because every model uses the same panel. The 1-km panel uses a local equidistant projection and Euclidean distance, i.e.\ ground meters.} The hourly main panel covers service hours 06:00--23:00 (18 steps per day, 360 steps in total). Mean hourly demand is 1.9 trips per station in each direction, 41\% of station-hours have zero departures, and the station distribution is heavily right-skewed: the ten busiest stations each exceed 20 departures per hour while a third of stations average below 0.5. The \emph{1-km panel}, built for the external benchmark of Section~\ref{sec:external}, follows the protocol of \citet{li2026cogclstm}: every record is attributed to the nearest of 336 station locations within 1~km (Euclidean), the full day is retained, and counts are formed at 15, 30 and 60 minutes over 21 days (12 December--1 January).

The observation window coincided with a sharp city-wide decline in travel: daily bike events fell from 1.2 million on 12 December to 0.3 million on 24 December before partially recovering. This non-stationarity is part of the forecasting problem and, as shown below, is one reason why metro flows matter at the network scale.

\subsection{Coupling diagnostics: Granger causality and distance-band elasticities}
\label{sec:coupling_results}

Before using metro flows as predictors, we verify that the two modes are coupled at short lags and establish over what distance the coupling extends. Bidirectional Granger causality tests \citep{granger1969investigating} are used to examine whether short-run predictive linkages exist between metro flows and bike usage. For each catchment $d\in\{50,100,200,500\}$\,m and each station with sufficient bike activity, the hourly series of metro flow (entries or exits) and bike trips (arrivals or departures) within service hours are $\log(1+y)$-transformed and first-differenced, which removes the level shift of the period and attenuates the shared intraday profile. We test whether metro flow Granger-causes bike flow and vice versa with lags of one to three hours, taking the smallest $F$-test $p$-value over the three lag lengths. The magnitude and spatial extent of the coupling are then quantified with station fixed-effect Poisson regressions estimated by pseudo-maximum likelihood (PPML) \citep{santossilva2006log}, which remain consistent under the over-dispersion of the counts. Every bike event is assigned to its nearest station and to one of seven non-overlapping rings $r\in\{$0--50, 50--100, 100--150, 150--200, 200--300, 300--400, 400--500$\}$\,m, and for each ring and direction pair (metro entries with bike arrivals, exits with departures) we estimate
\begin{equation}
\text{Bike}^{(r)}_{s,t}=\exp\!\left(\gamma_r\log\!\big(1+\text{Metro}_{s,t}\big)+\gamma'_r\log\!\big(1+\text{Metro}_{s,t-1}\big)+u_s+\zeta_{h(t)}+\psi_{d(t)}\right),
\label{eq:ppml}
\end{equation}
where $t$ indexes hourly intervals, $u_s$, $\zeta_{h(t)}$ and $\psi_{d(t)}$ are station, hour-of-day and date fixed effects and the lag absorbs the serial correlation of metro flow, so that $\gamma_r$ is the within-station elasticity of bike flow in ring $r$ with respect to contemporaneous metro flow, net of the daily profile and of the city-wide decline in travel; standard errors are clustered by station. The same model is estimated on cumulative disks 0--$R$, the catchments of the forecasting panels. The distance-band panels count every lock/unlock record, as in the 1-km protocol of Section~\ref{sec:external}, rather than the trip endpoints of the main panel (raw records outnumber trip endpoints by about 3.5 to 1), and the hourly sample retains only station-hours with complete 5-minute coverage; the elasticities are scale-free, so these choices affect the counts in Table~\ref{tab:rings}, which are therefore not comparable with those of the main panel, but not the comparison across distances.

Table~\ref{tab:granger} summarizes the Granger results. Metro flow Granger-causes bike flow at 97--100\% of the tested stations in every catchment, whereas the reverse direction is significant at 39--90\% of stations and is weakest in the smallest catchment. The predictive link therefore runs mainly from metro to bike.
Table~\ref{tab:rings} reports the hourly elasticities. The coupling is flat close to the station and decays beyond it: for arrivals $\gamma_r$ lies between 0.40 and 0.48 in every ring out to 300\,m, with overlapping confidence intervals, and falls to 0.23 at 300--400\,m and 0.13 at 400--500\,m; for departures the plateau is 0.37--0.39 out to 200\,m and the decline begins one ring earlier. Among the cumulative disks, 0--200\,m has the highest elasticity in both directions (IN 0.456, OUT 0.381), the 100\,m and 300\,m disks are statistically indistinguishable from it, and the 500\,m disk is 19--27\% lower because it increasingly contains bike activity unrelated to the metro. The coupling also appears directionally asymmetric: the entry$\to$arrival elasticity exceeds the exit$\to$departure elasticity in every ring, although the confidence intervals overlap. These diagnostics establish that metro flows are a short-run predictor of bike demand within about 300\,m of the station, with higher point estimates for arrivals than for departures; the forecasting experiments below therefore measure the predictive value of a relationship that is known to exist, over the catchment at which it is strongest.

\begin{table}[htbp]
\centering\small
\begin{threeparttable}
\caption{Station-level Granger causality tests between hourly metro and bike flows, by catchment.}
\label{tab:granger}
\begin{tabular}{lcccc}
\toprule
\textbf{Direction} & \textbf{Distance (m)} & \textbf{Stations} & \textbf{$p<0.05$ (\%)} & \textbf{Median $p$} \\
\midrule
FINT\_IN $\rightarrow$ ride\_in & 50 & 95 & 96.8 & $2.9\times10^{-6}$ \\
FINT\_IN $\rightarrow$ ride\_in & 100 & 200 & 99.0 & $6.6\times10^{-9}$ \\
FINT\_IN $\rightarrow$ ride\_in & 200 & 238 & 99.2 & $4.7\times10^{-11}$ \\
FINT\_IN $\rightarrow$ ride\_in & 500 & 239 & 100.0 & $5.3\times10^{-12}$ \\
\midrule
ride\_in $\rightarrow$ FINT\_IN & 50 & 95 & 38.9 & 0.12 \\
ride\_in $\rightarrow$ FINT\_IN & 100 & 200 & 64.0 & 0.023 \\
ride\_in $\rightarrow$ FINT\_IN & 200 & 238 & 79.0 & $1.5\times10^{-3}$ \\
ride\_in $\rightarrow$ FINT\_IN & 500 & 239 & 82.4 & $2.9\times10^{-4}$ \\
\midrule
FINT\_OUT $\rightarrow$ ride\_out & 50 & 95 & 96.8 & $5.8\times10^{-7}$ \\
FINT\_OUT $\rightarrow$ ride\_out & 100 & 200 & 99.0 & $3.2\times10^{-10}$ \\
FINT\_OUT $\rightarrow$ ride\_out & 200 & 238 & 100.0 & $7.8\times10^{-13}$ \\
FINT\_OUT $\rightarrow$ ride\_out & 500 & 239 & 100.0 & $2.9\times10^{-12}$ \\
\midrule
ride\_out $\rightarrow$ FINT\_OUT & 50 & 95 & 44.2 & 0.065 \\
ride\_out $\rightarrow$ FINT\_OUT & 100 & 200 & 67.5 & 0.018 \\
ride\_out $\rightarrow$ FINT\_OUT & 200 & 238 & 77.7 & $1.8\times10^{-3}$ \\
ride\_out $\rightarrow$ FINT\_OUT & 500 & 239 & 90.0 & $4.2\times10^{-5}$ \\
\bottomrule
\end{tabular}
\begin{tablenotes}\footnotesize
\item Notes: ride\_in and ride\_out denote bike-sharing arrivals and departures; FINT\_IN and FINT\_OUT denote metro passenger inflows and outflows. Distances are nominal (see footnote in Section~\ref{sec:data_prep}). Bike flows are trip endpoints, as in the main panel. Of the 342 stations, a station is tested if its mean hourly arrivals and departures both reach the larger of 0.2 and half the cross-station mean of hourly departures in that catchment. For each station, $p$ is the smallest $F$-test $p$-value over lags of 1--3 hours, without correction for the three lags.
\end{tablenotes}
\end{threeparttable}
\end{table}

\begin{table}[htbp]
\centering\small
\begin{threeparttable}
\caption{Elasticity of bike flow with respect to metro flow by distance band, hourly resolution.}
\label{tab:rings}
\begin{tabular}{lcccc}
\toprule
\textbf{Band (m)} & \textbf{IN $\gamma_r$ [95\% CI]} & \textbf{OUT $\gamma_r$ [95\% CI]} & \textbf{Nonzero \% IN/OUT} & \textbf{Mean IN/OUT} \\
\midrule
\multicolumn{5}{l}{\emph{Non-overlapping rings}} \\
0--50    & 0.399 [0.265, 0.533] & 0.366 [0.271, 0.461] & 55/72 & 1.7/2.7 \\
50--100  & 0.465 [0.383, 0.547] & 0.372 [0.297, 0.447] & 83/87 & 4.7/5.0 \\
100--150 & 0.450 [0.374, 0.525] & \textbf{0.391} [0.325, 0.457] & 93/93 & 6.8/6.4 \\
150--200 & \textbf{0.475} [0.385, 0.565] & \textbf{0.391} [0.313, 0.469] & 91/92 & 5.9/5.6 \\
200--300 & 0.429 [0.324, 0.535] & 0.279 [0.192, 0.365] & 97/97 & 8.6/8.6 \\
300--400 & 0.229 [0.139, 0.319] & 0.161 [0.098, 0.224] & 96/97 & 7.8/8.0 \\
400--500 & 0.132 [0.046, 0.218] & 0.105 [0.041, 0.169] & 96/96 & 7.6/7.9 \\
\midrule
\multicolumn{5}{l}{\emph{Cumulative disks}} \\
0--50    & 0.399 [0.265, 0.533] & 0.366 [0.271, 0.461] & 55/72   & 1.7/2.7 \\
0--100   & 0.447 [0.366, 0.529] & 0.371 [0.299, 0.444] & 88/93   & 6.4/7.7 \\
0--150   & 0.449 [0.380, 0.518] & \textbf{0.381} [0.316, 0.446] & 98/99   & 13.2/14.1 \\
0--200   & \textbf{0.456} [0.386, 0.525] & \textbf{0.381} [0.318, 0.444] & 100/100 & 19.1/19.7 \\
0--300   & 0.447 [0.382, 0.513] & 0.347 [0.285, 0.410] & 100/100 & 27.7/28.3 \\
0--400   & 0.409 [0.343, 0.475] & 0.311 [0.251, 0.370] & 100/100 & 35.5/36.3 \\
0--500   & 0.371 [0.305, 0.437] & 0.279 [0.221, 0.336] & 100/100 & 43.1/44.1 \\
\bottomrule
\end{tabular}
\begin{tablenotes}\footnotesize
\item Notes: PPML estimates of $\gamma_r$ in Eq.~(\ref{eq:ppml}) with station, hour-of-day and date fixed effects and a one-period metro lag; 95\% confidence intervals in brackets, standard errors clustered by station; all $p<0.01$. IN pairs metro entries with bike arrivals, OUT pairs metro exits with bike departures. Rings are non-overlapping and assigned by nearest station; disks are cumulative from the station. $N=20{,}526$ station-hours in every row: the ring panel is built from the 500\,m daily files, which contain 336 station names; metro flows are available only in 5-minute slots with at least one bike record within 500\,m, and the hourly sample keeps only station-hours with all twelve slots present and a preceding hour for the lag. Nonzero: share of station-hours with at least one bike event in the band; Mean: bike events per station-hour. Bold: largest estimate in each block (ties both in bold). Distances are nominal (see footnote in Section~\ref{sec:data_prep}).
\end{tablenotes}
\end{threeparttable}
\end{table}

\subsection{Spatial heterogeneity of station-area bike activity}
\label{sec:kde}

\begin{figure}[htbp]
  \centering
  \includegraphics[width=0.7\columnwidth]{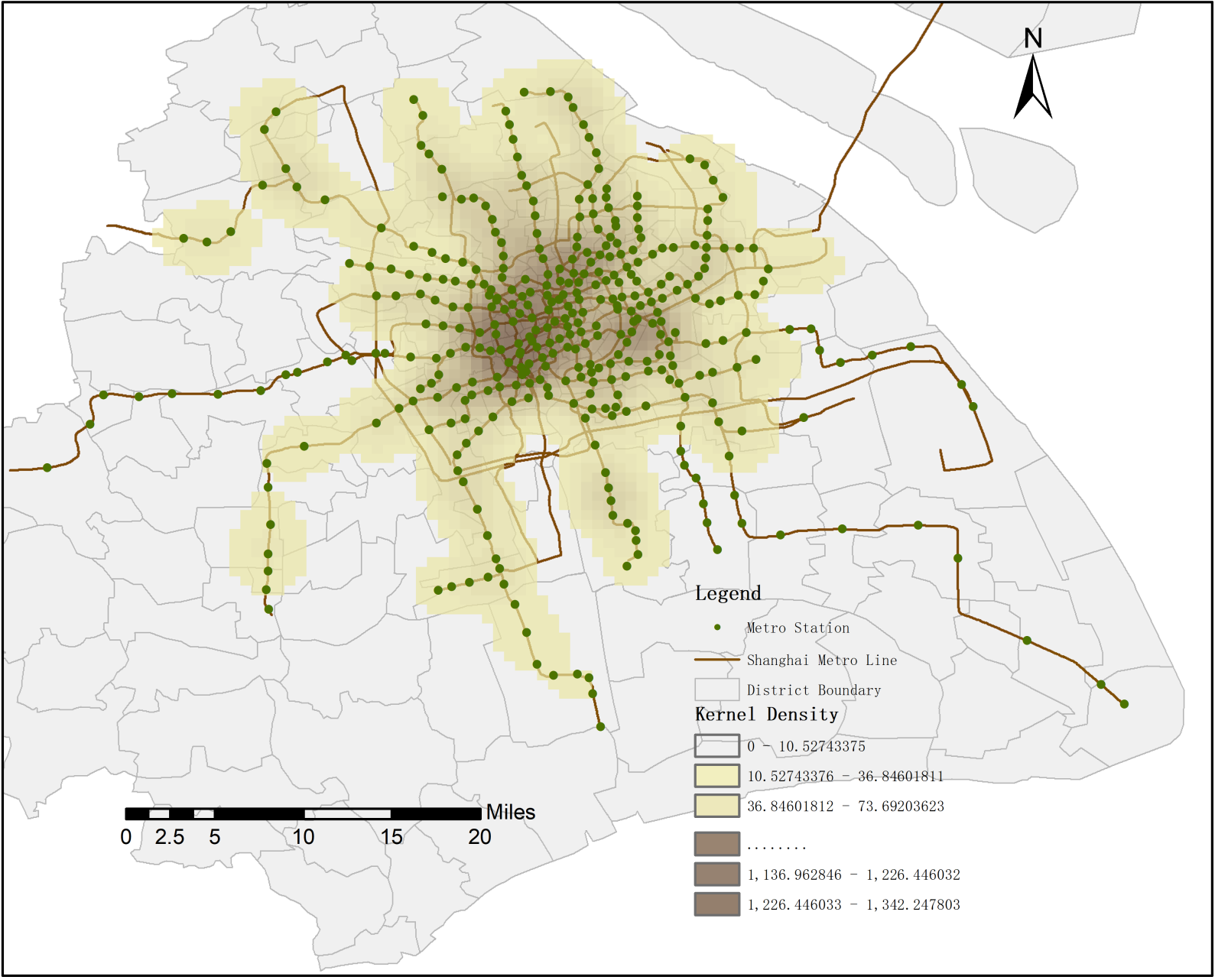}
  \caption{Kernel density of shared-bike lock/unlock events around metro stations in Shanghai.}
  \label{fig:kde_shanghai}
\end{figure}

Figure~\ref{fig:kde_shanghai} visualizes the spatial distribution of bike activity using kernel density estimation of lock and unlock events. Usage is highly uneven: densities are an order of magnitude higher inside the Inner Ring Road, peaking in the central districts where several metro lines intersect, and decline steadily outward, with stations beyond the Outer Ring and at line termini showing little surrounding activity. This heterogeneity is relevant to the forecasting comparison, since the gain from combining the statistical and deep forecasts (Section~\ref{sec:hybrid}) is largest at the high-volume hubs in the urban core.

\subsection{Forecasting task and evaluation protocol}
\label{sec:protocol}

The task is one-step-ahead prediction of \texttt{ride\_in} and \texttt{ride\_out} at all stations from the preceding $M$ hours of the four flow variables and five calendar features (hour and day-of-week encoded as sine/cosine pairs, plus a day-level activity indicator, the standardized logarithm of the previous day's network-wide bike volume). Results are reported at two scales. At the station level, every model produces forecasts for all 342 stations and errors are pooled over station--hours. At the network level, the 342 series are summed and models forecast the next hour's total, the quantity relevant to fleet sizing and to the question of whether metro throughput tracks day-level activity.

All models share one rolling-origin design with three folds \citep{tashman2000out,bergmeir2012use}. Fold $k$ trains on days 1 to $n_k$, uses the next two days for validation and the two days after that for testing; test windows are 26--27, 28--29 and 30--31 December, so that the last fold includes the weekend before New Year. The parameters of the statistical pipeline are selected separately in each fold on its training and validation days, and early stopping uses the validation days of the fold. The hyperparameters of the deep models are selected once and shared across folds (Appendix~\ref{app:hyper}), using the validation days of folds 1 and 3 at station level and of all three folds for the network total. Because the validation days of one fold are the test days of the preceding fold, this selection has seen the test days of fold 2 (station level) and of folds 1 and 2 (network total); the overlap favors the deep models and does not affect the statistical pipeline. Deep models are run with three random seeds per fold, so each configuration yields nine test evaluations, reported as mean $\pm$ standard deviation; this standard deviation reflects differences between folds as well as between seeds. Paired comparisons (e.g.\ with and without metro inputs) use Wilcoxon signed-rank tests \citep{wilcoxon1945individual} over the nine matched runs (one-sided, exact; the runs share three test windows and are not independent samples of future periods), or over stations for station-wise comparisons.

Accuracy is measured by mean absolute error (MAE), root mean squared error (RMSE) and $R^2$ on the original count scale. For the network total we also report the mean absolute percentage error and the bias ratio $\sum\hat y/\sum y$. For the external comparison we use the weighted absolute percentage error, $\mathrm{WMAPE}=\sum|y-\hat y|/\sum y$, which is scale-free and, unlike the mean absolute percentage error, remains defined at zero counts \citep{hyndman2006another}. Because counts are small, a large part of any error is sampling noise. As a reference we report the expected absolute error of a forecast equal to the mean if demand were Poisson with rate $\mu$, $\mathbb{E}|Y-\mu|$, which has the closed form given in Appendix~\ref{app:stat}. This Poisson reference is computed from the fitted rates. It is not a lower bound on MAE, because the MAE-optimal point forecast is the median and a forecast below the mean can score lower when $\mu$ is small; it indicates how much of the error of a mean forecast is attributable to count noise. For the probabilistic models we report the negative-binomial log score and the coverage of 90\% predictive intervals.

%=====================================================================
\section{Methodology}
\label{sec:method}
%=====================================================================

The models are chosen so that the three sources of improvement identified in Section~\ref{sec:intro} can be separated. All models receive the same input tensor and are evaluated under the protocol of Section~\ref{sec:protocol}; they differ only in the information they use (bike history alone, or bike and metro flows) and in the architecture that processes it. The statistical pipeline (Section~\ref{sec:stat}) is a four-stage multiplicative model of the conditional mean with a negative-binomial error model. It uses bike history only (no metro inputs), with cross-station information entering only through the shrinkage prior of stage (b); it has five scalar hyperparameters and no trainable weights, and each stage is stated with its probabilistic interpretation so that the meaning of every hyperparameter is fixed; derivations, the algorithm and the Poisson reference formula are given in Appendix~\ref{app:stat}. The deep models (Sections~\ref{sec:lstm}--\ref{sec:training}) are a full LSTM, which has no graph structure, and two graph architectures, ST-GCN and the cross-modal MBC-GCN, which propagate information along the metro network; each is trained with and without metro inputs. The hybrids (Section~\ref{sec:hybrid_design}) start a network from the statistical forecast, or average the two, to test whether the networks learn anything beyond it. Section~\ref{sec:problem} formulates the task and Table~\ref{tab:nomenclature} summarizes the notation.

\begin{table}[htbp]
\centering
\footnotesize
\setlength{\tabcolsep}{4pt}
\renewcommand{\arraystretch}{1.08}
\caption{Nomenclature.}
\label{tab:nomenclature}
\begin{tabularx}{\textwidth}{@{} l >{\raggedright\arraybackslash}X @{\hspace{1.2em}} l >{\raggedright\arraybackslash}X @{}}
\toprule
\textbf{Symbol} & \textbf{Description} & \textbf{Symbol} & \textbf{Description} \\
\midrule
$s=1,\dots,N$ & Metro station & $d=1,\dots,D$ & Service day \\
$h=1,\dots,H$ & Service hour within a day ($H=18$) & $t\equiv(d,h)$ & Global hourly time index \\
$y_{t,s}=y_{d,h,s}$ & Observed bike count (ride-in or ride-out) & $\mathrm{type}(d)$ & Day type: weekday or weekend \\
$\mathcal{D}_d$ & Earlier days of the same type as $d$ & $\mathcal{X}_{t-M+1:t}$ & Input tensor, $M$ steps, $N$ stations, $F$ features \\
$\mathcal{G}=(\mathcal{V},\mathcal{E},A)$ & Station graph, adjacency $A$ & $F$ & Feature dimension ($=9$: four flows, five calendar) \\
\midrule
$w_{d,i}$, $W_d$ & Recency weight of day $i$; effective number of days & $\lambda$, $\delta$ & Half-life of the weights (days); cold-start pseudo-days \\
$p_{d,h,s}$ & Recency-weighted intraday profile & $m_{d,h,s}=\ell_{d,s}\cdot g_{d,h}$ & Prior mean: station level $\times$ network shape \\
$b_{d,h,s}$ & Shrunk (empirical-Bayes) profile & $\kappa_{\mathrm{eb}}$, $\omega_{d,h,s}$ & Shrinkage strength; weight on the station's own profile \\
$r_{t,s}$ & Recent-level ratio at time $t$ & $\rho,\rho'$, $\beta$ & Pseudo-counts of the day and 3-h windows; 3-h weight \\
$\mu_{t+1,s}$ & Pipeline point forecast, $b_{t+1,s}\cdot r_{t,s}$ & $\phi$ & Negative-binomial dispersion \\
\midrule
$\mathbf{e}_t\in\mathbb{R}^{d_e}$, $\mathbf{X}^*$ & Time embedding; joint flow--time representation & $\tilde{\mathbf{L}}$, $T_k(\cdot)$, $\Theta_k$, $K$ & Rescaled Laplacian; Chebyshev polynomial, filter and order; $\boldsymbol{\Delta}$: degree matrix \\
$\hat{y}^{\mathrm{in}}_{t+1,s},\hat{y}^{\mathrm{out}}_{t+1,s}$ & Predicted arrivals and departures & $f(\cdot)$ & Forecasting map (any model) \\
\bottomrule
\end{tabularx}
\end{table}

\subsection{Problem formulation}
\label{sec:problem}

The forecasting task is defined over a network of $N$ metro stations observed hourly over $D$ service days of $H$ hours each, so that a global time index $t$ corresponds to a (day, hour) pair $(d,h)$. At each $t$, station $s$ is described by a feature vector $\mathbf{x}_{s,t}\in\mathbb{R}^{F}$ containing the shared-bike ride-in and ride-out counts, the metro entry and exit counts, and calendar indicators. The network state is $\mathbf{X}_t\in\mathbb{R}^{N\times F}$, and $M$ consecutive states are stacked into the tensor $\mathcal{X}_{t-M+1:t}\in\mathbb{R}^{M\times N\times F}$. The objective is to produce one-step-ahead forecasts of bike arrivals and departures at all stations simultaneously,
\begin{equation}
\left[\hat{y}^{\mathrm{in}}_{t+1},\;\hat{y}^{\mathrm{out}}_{t+1}\right]=f\!\left(\mathcal{X}_{t-M+1:t};\,\mathcal{G}\right),
\label{eq:problem}
\end{equation}
where $\mathcal{G}$ is the spatial graph used by the graph models and ignored by the others. In the \emph{AR} configuration the metro channels are removed from $\mathcal{X}$; in the \emph{Semi-AR} configuration they are retained. Arrivals and departures are treated identically, so the superscripts are dropped below and $y_{t,s}$ denotes either count.

\subsection{A calibrated statistical pipeline for station-level counts}
\label{sec:stat}

\begin{figure}[htbp]
\centering
\begin{tikzpicture}[
  font=\footnotesize,
  stage/.style={rectangle, draw=gray!70, fill=gray!8, rounded corners=3pt, text width=2.55cm, minimum height=1.4cm, align=center},
  data/.style={rectangle, draw=blue!60, fill=blue!10, rounded corners=3pt, text width=2.1cm, minimum height=1.4cm, align=center},
  outbox/.style={rectangle, draw=red!60, fill=red!8, rounded corners=3pt, text width=2.5cm, minimum height=1.4cm, align=center},
  arr/.style={-{Stealth[length=2mm]}, thick, draw=gray!70},
  lab/.style={font=\scriptsize, text=gray!90, align=center}
]
\node[data] (hist) {Earlier days\\$\{y_{i,h,s}\}_{i<d}$};
\node[stage, right=0.45cm of hist] (a) {\textbf{(a)} Recency-weighted profile\\$p_{d,h,s}$};
\node[stage, right=0.45cm of a] (b) {\textbf{(b)} Empirical-Bayes shrinkage\\$b_{d,h,s}$};
\node[stage, right=0.45cm of b] (c) {\textbf{(c)} Recent-level ratio\\$\mu_{t+1,s}=b\cdot r_{t,s}$};
\node[outbox, right=0.45cm of c] (d) {\textbf{(d)} Negative binomial\\$Y_{t+1,s}\sim\mathrm{NB}(\mu,\phi)$};
\node[data, below=0.55cm of b, text width=3.0cm] (net) {Network shape $g_{d,h}$,\\station level $\ell_{d,s}$ (days $<d$)};
\node[data, below=0.55cm of c, text width=3.0cm] (rec) {Same-day history\\$\{y_{\tau,s}\}_{\tau\le t}$};
\draw[arr] (hist) -- (a); \draw[arr] (a) -- (b); \draw[arr] (b) -- (c); \draw[arr] (c) -- (d);
\draw[arr] (net) -- (b); \draw[arr] (rec) -- (c);
\node[lab, above=0.05cm of a] {$\lambda,\ \delta$};
\node[lab, above=0.05cm of b] {$\kappa_{\mathrm{eb}}$};
\node[lab, above=0.05cm of c] {$\rho,\ \rho',\ \beta$};
\node[lab, above=0.05cm of d] {$\phi$};
\end{tikzpicture}
\caption{The four stages of the statistical pipeline and the information each uses. Stages (a)--(b) use only days before the target day; stage (c) adds the current day up to the forecast origin $t$. Hyperparameters are shown above each stage.}
\label{fig:pipeline_stat}
\end{figure}
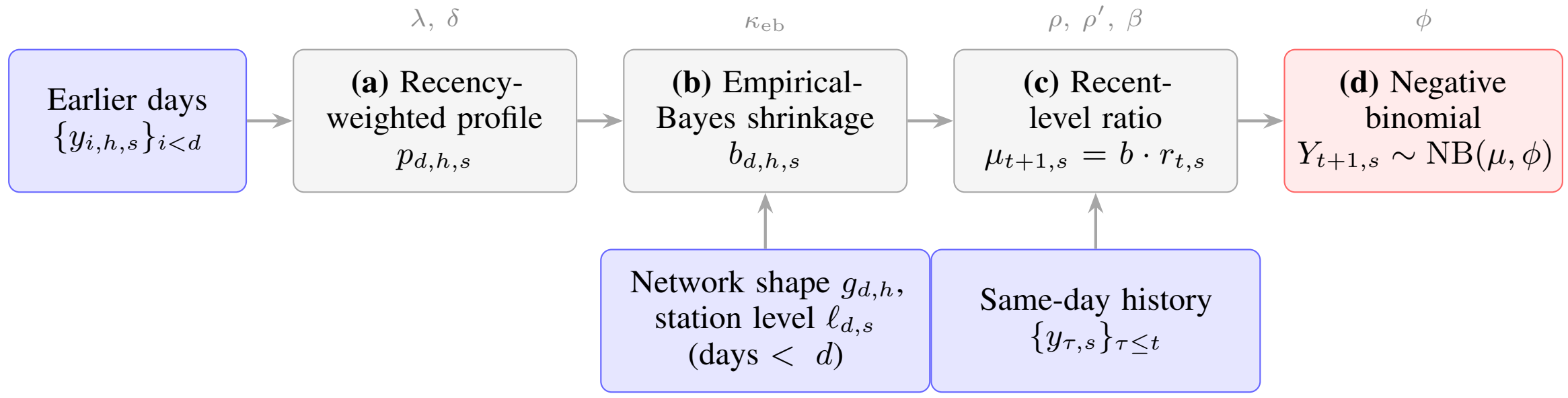

The design of the pipeline follows from three properties of the panel. The targets are small, zero-inflated counts (41\% of station--hours are zero), so the pipeline models them with a count distribution, and part of any forecast error is Poisson noise. Demand has a strong intraday periodicity, and station levels differ by two orders of magnitude; the pipeline assumes that stations share an intraday shape and shrinks each station's own profile toward it. Finally, the overall level of demand changes from day to day over the study window, so the profile is rescaled by the level observed on the current day. The pipeline is therefore a multiplicative model of the conditional mean with a negative-binomial observation model,
\begin{equation}
Y_{t+1,s}\sim\mathrm{NB}\!\left(\mu_{t+1,s},\phi\right),\qquad
\mu_{t+1,s}=\underbrace{b_{d(t+1),h(t+1),s}}_{\text{profile: (a)+(b)}}\times\underbrace{r_{t,s}}_{\text{recent level: (c)}},
\label{eq:pipeline}
\end{equation}
where $Y_{t+1,s}$ is the count at station $s$ in hour $t+1$, $\mu_{t+1,s}$ is its conditional mean, $\phi$ is the dispersion parameter of the negative-binomial (NB) distribution, and $d(t+1)$ and $h(t+1)$ denote the service day and the hour of day of step $t+1$. The profile $b_{d,h,s}$ carries the shape of the day, estimated from earlier days and shrunk toward the network-wide shape; the ratio $r_{t,s}$ carries the level of the current day, estimated from the most recent observations. The four components (a)--(d) are added cumulatively (Figure~\ref{fig:pipeline_stat}). Each is a one- or two-parameter step with a closed-form Bayesian interpretation, and the pipeline is causal (the forecast for $t+1$ uses only observations up to $t$, and the profile for day $d$ only days before $d$), closed-form (no trainable weights, five scalar hyperparameters), decomposable (every forecast factorizes as in Eq.~(\ref{eq:pipeline})) and probabilistic.

\paragraph{(a) Recency-weighted intraday profile}
Let $\mathcal{D}_d=\{i<d:\mathrm{type}(i)=\mathrm{type}(d)\}$ be the set of earlier days of the same type (weekday or weekend) as the target day $d$. The profile is the weighted mean of the counts observed in the same hour on those days,
\begin{equation}
p_{d,h,s}=\frac{\sum_{i\in\mathcal{D}_d}w_{d,i}\,y_{i,h,s}}{W_d},\qquad
w_{d,i}=2^{-(d-i-1)/\lambda},\qquad
W_d=\sum_{i\in\mathcal{D}_d}w_{d,i},
\label{eq:profile}
\end{equation}
where $y_{i,h,s}$ is the count observed at station $s$ in hour $h$ of day $i$, $w_{d,i}$ is the weight of day $i$ when forecasting day $d$, $\lambda>0$ is the half-life of the weights in days, and $W_d$, the sum of the weights, is the effective number of days behind the estimate; $W_d$ is used in stage (b) to quantify how much the estimate should be trusted. Two limiting cases are familiar estimators: $\lambda\to\infty$ gives the historical mean of the same hour and day type (\texttt{hist\_mean}) and $\lambda\to0$ the most recent day of the same type; an intermediate $\lambda$ is exponential smoothing across days within each hour-of-day cell with smoothing constant $\alpha=1-2^{-1/\lambda}$, the seasonal component of a multiplicative Holt--Winters scheme without trend \citep{winters1960forecasting,hyndman2008forecasting}. For Poisson counts with mean $\theta$, the variance of $p_{d,h,s}$ is of order $\theta/W_d$, so a half-life of one or two days yields a coefficient of variation above one wherever $\theta<1$; stage (b) is designed to reduce this noise. On the first days of each type, when $W_d$ is small, $p_{d,h,s}$ is blended with the all-day profile $p^{\mathrm{all}}_{d,h,s}$, obtained from Eq.~(\ref{eq:profile}) with all earlier days, through $\delta\ge0$ pseudo-days: $\tilde p_{d,h,s}=(W_d\cdot p_{d,h,s}+\delta\cdot p^{\mathrm{all}}_{d,h,s})/(W_d+\delta)$ and $\tilde W_d=W_d+\delta$ replace $p_{d,h,s}$ and $W_d$ in the following stages. The blend is negligible once $W_d\gg\delta$, and $\delta=0$ disables it.

\paragraph{(b) Empirical-Bayes shrinkage toward the network shape}
Station profiles estimated from a handful of days are noisy at every station and unusable at low-volume ones. Each profile is therefore shrunk toward a prior mean $m_{d,h,s}$ that encodes the one well-supported piece of cross-station structure, a shared intraday shape with station-specific levels:
\begin{equation}
m_{d,h,s}=\ell_{d,s}\cdot g_{d,h},\qquad
\ell_{d,s}=\frac1H\sum_{h'}p_{d,h',s},\qquad
g_{d,h}=\frac{\sum_{s'}p_{d,h,s'}}{\frac1H\sum_{h'}\sum_{s'}p_{d,h',s'}},
\label{eq:prior}
\end{equation}
where $\ell_{d,s}$ is the recency-weighted mean hourly count of station $s$, $g_{d,h}$ is the network-wide profile normalized to unit daily mean, $H=18$ is the number of service hours per day, and $h'$ and $s'$ index hours and stations. Both quantities are causal, and $\sum_h m_{d,h,s}=\sum_h p_{d,h,s}$: the prior preserves each station's daily total and only redistributes it across hours. Let $\theta_{h,s}$ denote the unknown mean count of station $s$ in hour $h$ of the target day. With the prior $\theta_{h,s}\sim\mathrm{Gamma}(\kappa_{\mathrm{eb}},\kappa_{\mathrm{eb}}/m_{d,h,s})$, whose mean is $m_{d,h,s}$ and whose shape $\kappa_{\mathrm{eb}}>0$ is the prior precision, the posterior mean under the recency-weighted Poisson likelihood is (Appendix~\ref{app:stat_eb})
\begin{equation}
b_{d,h,s}=\frac{\kappa_{\mathrm{eb}}+W_d\cdot p_{d,h,s}}{\kappa_{\mathrm{eb}}/m_{d,h,s}+W_d}
=\omega_{d,h,s}\cdot p_{d,h,s}+(1-\omega_{d,h,s})\cdot m_{d,h,s},\qquad
\omega_{d,h,s}=\frac{W_d}{W_d+\kappa_{\mathrm{eb}}/m_{d,h,s}},
\label{eq:eb}
\end{equation}
which is a weighted average of the station's own profile and the prior mean with weight $\omega_{d,h,s}\in(0,1)$ on the station, so that $b_{d,h,s}$ always lies between the two. The prior is worth $\kappa_{\mathrm{eb}}/m_{d,h,s}$ pseudo-days, inversely proportional to the expected count, which allows a single $\kappa_{\mathrm{eb}}$ to be used for the whole network: with $\kappa_{\mathrm{eb}}=10$ and $W_d=3$, a hub with $m=20$ keeps 86\% of its own profile ($\omega=0.86$), whereas a tail station with $m=0.2$ keeps 6\%, so that shrinkage is strongest where the Poisson noise of stage (a) is largest. The estimator is the count analog of the James--Stein rule \citep{james1961estimation,efron1973stein} and the Gamma--Poisson shrinkage of small-area estimation \citep{robbins1992empirical,clayton1987empirical}; the prior mean is estimated from the data and the prior precision $\kappa_{\mathrm{eb}}$ by validation. Stage (b) is the only place where the pipeline uses cross-station information, and its only assumption is that stations share an intraday shape.

\paragraph{(c) Recent-level ratio}
The profile assumes that the target day resembles earlier days of the same type, an assumption that the study period violates. The level of the current day is therefore treated as a latent multiplicative factor $r$ and estimated from a causal window $\mathcal{W}(t)$ of hours $\tau\le t$: if the counts in the window are Poisson with means $r\cdot b_{\tau,s}$ and $r$ has the conjugate prior $\mathrm{Gamma}(\rho,\rho)$, centered on one with pseudo-count $\rho>0$, the posterior mean of $r$ is a pseudo-count ratio (Appendix~\ref{app:stat_ratio}). Two windows are used, the previous service day and the previous three hours, with pseudo-counts $\rho$ and $\rho'=3\rho/H$, and their ratios are pooled log-linearly:
\begin{equation}
\hat r_{\mathrm{day}}(t,s)=\frac{\sum_{\tau\in\mathcal{W}_{\mathrm{day}}(t)} y_{\tau,s}+\rho}{\sum_{\tau\in\mathcal{W}_{\mathrm{day}}(t)} b_{\tau,s}+\rho},\qquad
\hat r_{3\mathrm{h}}(t,s)=\frac{\sum_{\tau\in\mathcal{W}_{3\mathrm{h}}(t)} y_{\tau,s}+\rho'}{\sum_{\tau\in\mathcal{W}_{3\mathrm{h}}(t)} b_{\tau,s}+\rho'},
\label{eq:ratio_window}
\end{equation}
\begin{equation}
r_{t,s}=\hat r_{\mathrm{day}}(t,s)^{\,1-\beta}\cdot\hat r_{3\mathrm{h}}(t,s)^{\,\beta},\qquad
\mu_{t+1,s}=b_{d(t+1),\,h(t+1),\,s}\cdot r_{t,s},
\label{eq:ratio}
\end{equation}
where $\mathcal{W}_{\mathrm{day}}(t)=\{t-H+1,\dots,t\}$, $\mathcal{W}_{3\mathrm{h}}(t)=\{t-2,t-1,t\}$, $b_{\tau,s}$ is the shrunk profile of stage (b) for hour $\tau$, and $\beta\in[0,1]$ is the weight of the three-hour window. The pseudo-count anchors the ratio at one where the window contains few expected trips, so that a quiet hour at a low-volume station does not forecast zero for the rest of the day, whereas at a hub, where $\sum_\tau b_{\tau,s}$ runs to hundreds, it is immaterial. The day window tracks level shifts that persist over days (the city-wide decline, a holiday) with little variance; the three-hour window reacts to intraday shocks at the cost of more noise; $\beta$ interpolates between them. The geometric pool is the natural average of a multiplicative factor, since a doubling in one window and a halving in the other cancel exactly. Because every $b_{\tau,s}$ in the window depends only on days before the day of $\tau$, the denominators are known at $t$ and the ratio is causal. Applied to the network total, this step allows the model to follow the aggregate decline in travel.

\paragraph{(d) Negative-binomial predictive distribution}
Conditional on $\mu_{t+1,s}$, hourly counts are over-dispersed, so the predictive distribution is negative binomial in the NB2 parameterization, $\mathrm{Var}[Y]=\mu+\mu^2/\phi$ \citep{cameron2013regression}, which tends to the Poisson as $\phi\to\infty$. Because over-dispersion scales with volume, $\phi$ is estimated by maximum likelihood separately for the low-, mid- and high-volume station groups (training mean below 0.5, 0.5--2 and above 2 trips per hour) with $\mu$ held fixed at the pipeline forecast, which calibrates the spread without changing the point forecast. The distribution yields 90\% predictive intervals and the logarithmic score, a strictly proper scoring rule \citep{gneiting2007strictly} on which the hybrids of Section~\ref{sec:hybrid_design} are trained and compared. The point forecast reported is the mean $\mu$. For right-skewed counts the median lies below the mean, so a forecaster optimized for absolute error would report a smaller, downward-biased value and obtain a lower MAE; for this reason the bias ratio, RMSE and the log score are reported alongside MAE.

\paragraph{Hyperparameter selection}
The five hyperparameters $(\lambda,\delta,\kappa_{\mathrm{eb}},\rho,\beta)$ are selected by minimizing the Poisson deviance of the one-step forecasts,
\begin{equation}
\mathrm{Dev}=2\sum_{t,s}\Big[y_{t+1,s}\log\frac{y_{t+1,s}}{\mu_{t+1,s}}-(y_{t+1,s}-\mu_{t+1,s})\Big],
\label{eq:deviance}
\end{equation}
where the sum runs over all station--hours of the training and validation days of a fold and the term for $y_{t+1,s}=0$ equals $2\mu_{t+1,s}$. The deviance is twice the log-likelihood ratio between a saturated Poisson model, which fits each count exactly, and the forecast $\mu$; it is zero when every forecast equals its outcome and grows with every miss. It is used for three reasons. It equals the Poisson log score up to a constant, so that the parameters are selected on the scale on which the forecasts are later evaluated. Each term is scaled by the count itself, so that a 10\% miss at a station with 20 trips per hour costs more than the same relative miss at a station with 2, but linearly rather than quadratically as under squared error; neither the busiest stations nor the long tail of near-empty ones therefore dominates the selection. It is also defined for the zeros that make up a large share of the panel, where relative errors are not. The search is exhaustive: for each fold, every combination on the grid in Appendix~\ref{app:grid} is run and the combination with the smallest deviance is kept; test days are never used. Because every forecast uses only observations before its target, the deviance on training days is already an out-of-sample one-step criterion. The dispersion $\phi$ is fitted afterwards by maximum likelihood with $\mu$ held at the selected forecast. Table~\ref{tab:stat_hyper} lists the hyperparameters, their limiting cases and the values selected.

\paragraph{Computational cost and reference forecasts}
Each hyperparameter combination requires a single pass over the panel, so the cost of the pipeline grows linearly with the number of days $D$, hours per day $H$ and stations $N$ ($20\times18\times342\approx1.2\times10^5$ station--hours in the full panel); it runs in under a minute including the grid search, is deterministic, and its forecast at every station--hour decomposes into the profile, the shrinkage weight and the ratio that produced it (Algorithm~\ref{alg:pipeline} in Appendix~\ref{app:stat}). Because the stages are cumulative, the results report the pipeline truncated after each stage, labeled (a), (a+b), (a--c) and (a--d), where (a--c) is the point forecast. Three naive forecasts serve as references: the historical mean of the same hour and day type (\texttt{hist\_mean}), the same hour on the previous day (\texttt{seasonal\_naive}) and the previous hour (\texttt{persistence}).

\begin{table}[htbp]
\centering\small
\renewcommand{\arraystretch}{1.12}
\caption{Hyperparameters of the statistical pipeline, limiting cases, and values selected on the main 200-m panel and on the 1-km panel of the external comparison (ranges span the three folds).}
\label{tab:stat_hyper}
\begin{tabularx}{\textwidth}{@{} l l >{\raggedright\arraybackslash}X l l @{}}
\toprule
Symbol & Stage & Role and limiting cases & Main panel & 1-km panel \\
\midrule
$\lambda$ & (a) & Half-life of day weights; $\infty$: same-hour historical mean; $0$: most recent same-type day & 2--5 d & 2--3 d \\
$\delta$ & (a) & Cold-start pseudo-days toward the all-day profile; $0$ disables & 0--2 & 0 \\
$\kappa_{\mathrm{eb}}$ & (b) & Prior precision; prior worth $\kappa_{\mathrm{eb}}/m$ days at expected count $m$; $0$: no shrinkage; $\infty$: level $\times$ network shape & 10 & 30 \\
$\rho$ & (c) & Pseudo-count of the day window ($\rho'=3\rho/H$); $\infty$: no level correction & 5--20 & 50 \\
$\beta$ & (c) & Weight of the 3-h window; $0$: day window only; $1$: 3-h only & 0.25--0.5 & 0.75 \\
$\phi$ & (d) & NB2 dispersion, fitted by maximum likelihood per volume group; $\infty$: Poisson & fitted & fitted \\
\bottomrule
\end{tabularx}
\end{table}

\subsection{Temporal baseline: LSTM}
\label{sec:lstm}

LSTM networks \citep{hochreiter1997long} serve as the sequential baseline, given their established capacity to capture long-range temporal dependencies and periodic travel patterns in traffic time series \citep{ma2015long}. At each step $t$ of the input window the unit updates its cell state $\mathbf{c}_t$ and hidden state $\mathbf{q}_t$ through three gating operations:
\begin{equation}
\begin{aligned}
i_t &= \sigma\!\left(\mathbf{U}_i\cdot[\mathbf{q}_{t-1},\mathbf{x}_t]+\mathbf{v}_i\right),\quad
f_t = \sigma\!\left(\mathbf{U}_f\cdot[\mathbf{q}_{t-1},\mathbf{x}_t]+\mathbf{v}_f\right),\quad
o_t = \sigma\!\left(\mathbf{U}_o\cdot[\mathbf{q}_{t-1},\mathbf{x}_t]+\mathbf{v}_o\right),\\
\tilde{\mathbf{c}}_t &= \tanh\!\left(\mathbf{U}_c\cdot[\mathbf{q}_{t-1},\mathbf{x}_t]+\mathbf{v}_c\right),\quad
\mathbf{c}_t = f_t\odot \mathbf{c}_{t-1}+i_t\odot\tilde{\mathbf{c}}_t,\quad
\mathbf{q}_t = o_t\odot\tanh(\mathbf{c}_t),
\end{aligned}
\label{eq:lstm}
\end{equation}
where $\mathbf{x}_t$ is the input vector at step $t$, $[\cdot,\cdot]$ denotes concatenation, $i_t$, $f_t$ and $o_t$ are the input, forget and output gates, $\mathbf{U}_\bullet$ and $\mathbf{v}_\bullet$ are learnable weights and biases, $\sigma(\cdot)$ is the logistic function, $\tilde{\mathbf{c}}_t$ is the candidate cell state and $\odot$ denotes element-wise multiplication. In the full LSTM used throughout, the $N\times F$ station features of each step are flattened into one vector $\mathbf{x}_t\in\mathbb{R}^{NF}$, so that every output can depend on the history of every station through the fully connected input and output maps.

\subsection{Spatial graph construction}
\label{sec:graph}

The graph models represent the metro network as an undirected weighted graph $\mathcal{G}=(\mathcal{V},\mathcal{E},A)$ \citep{shuman2013emerging}, where the nodes $\mathcal{V}$ are the $N$ stations, each carrying the feature vector of Section~\ref{sec:problem}, $\mathcal{E}$ is the set of edges and $A\in\mathbb{R}^{N\times N}$ the weighted adjacency matrix. Edge weights represent inter-station proximity through a thresholded Gaussian kernel on station-to-station distance,
\begin{equation}
A_{ss'}=
\begin{cases}
\exp\!\left(-d_{ss'}^{2}/2\sigma_{\mathrm{g}}^{2}\right) & \text{if } d_{ss'}\le\kappa_{\mathrm{g}},\\
0 & \text{otherwise},
\end{cases}
\label{eq:adj}
\end{equation}
where $d_{ss'}$ is the distance between stations $s$ and $s'$ (nominal Web-Mercator distance, as in Section~\ref{sec:data_prep}), $\sigma_{\mathrm{g}}=1$\,km is the kernel width and $\kappa_{\mathrm{g}}=2$\,km the cut-off distance. The cut-off is set so that each station is connected to its neighboring stations along the line and at transfer nodes (2.9 non-zero neighbors per station on average in the Shanghai network), which keeps the graph sparse. The convolutions use the symmetrically normalized adjacency with self-loops, $\tilde A_{\mathrm{norm}}=\boldsymbol{\Delta}^{-1/2}(A+I)\boldsymbol{\Delta}^{-1/2}$, where $I$ is the identity matrix and $\boldsymbol{\Delta}$ is the diagonal degree matrix with $\Delta_{ss}=\sum_{s'}(A+I)_{ss'}$; the normalization ensures numerical stability during iterative convolutions.

\subsection{Graph architectures: ST-GCN and MBC-GCN}
\label{sec:mbcgcn}

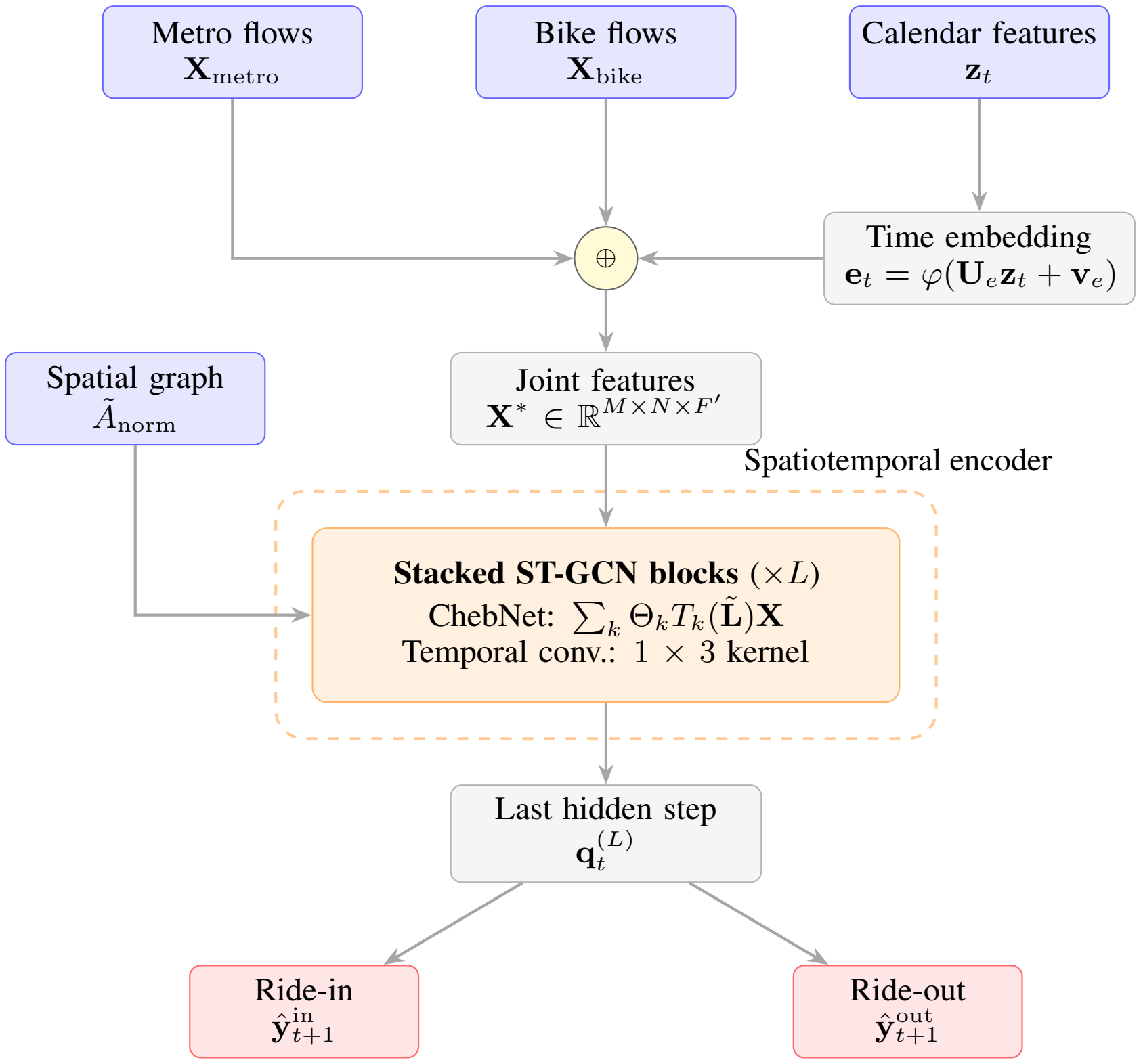
\begin{figure}[htbp]
\centering
\begin{tikzpicture}[
    node distance=0.8cm and 1.1cm,
    inputbox/.style={rectangle, draw=blue!60, fill=blue!10, text width=2.3cm, minimum height=0.9cm, align=center, rounded corners=3pt, font=\footnotesize},
    processbox/.style={rectangle, draw=gray!60, fill=gray!8, text width=2.8cm, minimum height=0.9cm, align=center, rounded corners=3pt, font=\footnotesize},
    modulebox/.style={rectangle, draw=orange!60, fill=orange!12, text width=3cm, minimum height=1cm, align=center, rounded corners=4pt, font=\footnotesize},
    outputbox/.style={rectangle, draw=red!60, fill=red!10, text width=2cm, minimum height=0.9cm, align=center, rounded corners=3pt, font=\footnotesize},
    arrow/.style={-{Stealth[length=2mm]}, thick, draw=gray!70},
    concat/.style={circle, draw=black!60, fill=yellow!20, minimum size=0.5cm, font=\scriptsize},
]
\node[inputbox] (metro) {Metro flows\\$\mathbf{X}_{\mathrm{metro}}$};
\node[inputbox, right=of metro] (bike) {Bike flows\\$\mathbf{X}_{\mathrm{bike}}$};
\node[inputbox, right=of bike] (time) {Calendar features\\$\mathbf{z}_t$};
\node[processbox, below=1.1cm of time] (embed) {Time embedding\\$\mathbf{e}_t=\varphi(\mathbf{U}_e\mathbf{z}_t+\mathbf{v}_e)$};
\draw[arrow] (time) -- (embed);
\node[concat] (cat1) at (bike |- embed) {$\oplus$};
\draw[arrow] (metro) |- (cat1); \draw[arrow] (bike) -- (cat1); \draw[arrow] (embed) -- (cat1);
\node[processbox, below=0.6cm of cat1] (joint) {Joint features\\$\mathbf{X}^*\in\mathbb{R}^{M\times N\times F'}$};
\draw[arrow] (cat1) -- (joint);
\node[inputbox, left=1.8cm of joint] (graph) {Spatial graph\\$\tilde A_{\mathrm{norm}}$};
\node[modulebox, below=0.8cm of joint, text width=5.5cm, minimum height=1.7cm] (stgcn) {
    \textbf{Stacked ST-GCN blocks} ($\times L$)\\[2pt]
    ChebNet: $\sum_k\Theta_kT_k(\tilde{\mathbf{L}})\mathbf{X}$\\
    Temporal conv.: $1\times3$ kernel};
\draw[arrow] (joint) -- (stgcn); \draw[arrow] (graph) |- (stgcn);
\node[processbox, below=0.8cm of stgcn] (attended) {Last hidden step\\$\mathbf{q}^{(L)}_{t}$};
\draw[arrow] (stgcn) -- (attended);
\node[outputbox, below left=0.8cm and 0.3cm of attended] (out_in) {Ride-in\\$\hat{\mathbf{y}}^{\mathrm{in}}_{t+1}$};
\node[outputbox, below right=0.8cm and 0.3cm of attended] (out_out) {Ride-out\\$\hat{\mathbf{y}}^{\mathrm{out}}_{t+1}$};
\draw[arrow] (attended) -- (out_in); \draw[arrow] (attended) -- (out_out);
\begin{scope}[on background layer]
    \node[draw=orange!40, dashed, thick, rounded corners=8pt, fit=(stgcn), inner sep=10pt,
          label={[font=\footnotesize]above right:Spatiotemporal encoder}] {};
\end{scope}
\end{tikzpicture}
\caption{Architecture of MBC-GCN. Metro and bike flows are concatenated with a learned calendar embedding and processed through stacked spatiotemporal graph-convolutional blocks with Chebyshev spectral filtering; the last hidden step is passed to two linear heads that produce joint predictions of ride-in and ride-out demand.}
\label{fig:mtgcn_architecture}
\end{figure}

Both graph models process an input representation $\mathbf{X}^*\in\mathbb{R}^{M\times N\times F'}$, derived from the input tensor $\mathcal{X}$ of Section~\ref{sec:problem} as described below, through $L$ stacked spatiotemporal blocks. Each block combines a $1\times3$ temporal convolution with a Chebyshev spectral graph convolution \citep{defferrard2016convolutional}, which aggregates spatial information up to $K$ hops ($K$ is tuned with the other hyperparameters, Section~\ref{sec:training}):
\begin{equation}
\mathrm{GCN}(\mathbf{X})=\sum_{k=0}^{K-1}\Theta_k\,T_k\!\left(\tilde{\mathbf{L}}\right)\mathbf{X},
\label{eq:chebconv}
\end{equation}
where $\mathbf{X}$ denotes the node features entering the convolution, $\tilde{\mathbf{L}}=2\mathbf{L}/\xi_{\max}-I$ is the rescaled graph Laplacian, $\mathbf{L}=I-\boldsymbol{\Delta}_{A}^{-1/2}A\boldsymbol{\Delta}_{A}^{-1/2}$ is the normalized Laplacian with $\boldsymbol{\Delta}_{A}$ the degree matrix of $A$ and $\xi_{\max}$ the largest eigenvalue of $\mathbf{L}$, $T_k(\cdot)$ is the Chebyshev polynomial of order $k$, given by the recursion $T_0=I$, $T_1=\tilde{\mathbf{L}}$, $T_k=2\tilde{\mathbf{L}}T_{k-1}-T_{k-2}$, and $\Theta_k$ are the learnable filter coefficients. Within block $l=1,\dots,L$, residual connections, layer normalization and dropout stabilize training on the 20-day window:
\begin{equation}
\begin{aligned}
\mathbf{Z}_t^{(l)}&=\mathrm{Dropout}\!\left(\mathrm{ReLU}\!\left(\mathrm{Conv}_{1\times3}\!\left(\mathbf{Q}^{(l-1)}\right)\right)\right),\quad
\mathbf{Z}_s^{(l)}=\mathrm{Dropout}\!\left(\mathrm{ReLU}\!\left(\mathrm{GCN}\!\left(\mathbf{Z}_t^{(l)}\right)\right)\right),\\
\mathbf{Q}^{(l)}&=\mathrm{LayerNorm}\!\left(\mathbf{Z}_s^{(l)}+\mathbf{Z}_t^{(l)}\right),\qquad \mathbf{Q}^{(0)}=\mathbf{X}^*,
\end{aligned}
\label{eq:stgcn_block}
\end{equation}
where $\mathbf{Z}_t^{(l)}$ and $\mathbf{Z}_s^{(l)}$ are the temporal and spatial feature maps of block $l$ and $\mathbf{Q}^{(l)}$ is its output. The two models differ in the construction of $\mathbf{X}^*$. ST-GCN \citep{yu2018spatio} feeds all $F$ channels of $\mathcal{X}$, flows and calendar features alike, into the blocks, so that $\mathbf{X}^*=\mathcal{X}$. MBC-GCN (Figure~\ref{fig:mtgcn_architecture}), the cross-modal graph network, differs in two respects. First, the five calendar features of Section~\ref{sec:protocol}, $\mathbf{z}_t=[\sin(\mathrm{hour}),\cos(\mathrm{hour}),\sin(\mathrm{dow}),\cos(\mathrm{dow}),\eta_t]^{\top}$ with $\mathrm{dow}$ the day of week and $\eta_t$ the day-level activity indicator, are projected through a learnable embedding $\mathbf{e}_t=\varphi(\mathbf{U}_e\mathbf{z}_t+\mathbf{v}_e)\in\mathbb{R}^{d_e}$, with $\varphi(\cdot)$ a nonlinear activation, $\mathbf{U}_e$ and $\mathbf{v}_e$ learnable weights and biases and $d_e$ the embedding dimension, before concatenation with the flow channels $\mathbf{X}_{\mathrm{flow}}$ of $\mathcal{X}$, $\mathbf{X}^*=[\mathbf{X}_{\mathrm{flow}},\mathbf{e}_t]$ with $F'=4+d_e$, so that peak/off-peak and weekday/weekend contrasts are learned rather than imposed. Second, metro flows enter as additional channels of the shared backbone, so that the convolution operates on the joint metro--bike state of each neighborhood. After the $L$-th block, the hidden state of the last step of $\mathbf{Q}^{(L)}$, denoted $\mathbf{q}^{(L)}_{t}$, is passed to two linear heads that produce $\hat y^{\mathrm{in}}_{t+1}$ and $\hat y^{\mathrm{out}}_{t+1}$.

\subsection{Training and tuning of the deep models}
\label{sec:training}

All deep models receive the same input tensor with $F=9$ channels (four flows, five calendar features); flow channels are $\log(1+y)$-transformed and standardized on training days, and predictions are mapped back to counts. Each architecture is trained in the AR and Semi-AR configurations, giving six configurations. Training minimizes mean squared error on the transformed scale with AdamW \citep{loshchilov2019decoupled}, five warm-up epochs followed by cosine decay, gradient clipping, up to 200 epochs and early stopping with patience 20 on validation loss (300 epochs and patience 30 for the network-total series). Hyperparameters (hidden width, depth, dropout, learning rate, weight decay, batch size, sequence length and, for the graph models, Chebyshev order) are tuned per architecture with Optuna \citep{akiba2019optuna} on validation folds; selected values are listed in Appendix~\ref{app:hyper}. The deep models and the statistical pipeline are evaluated on the same held-out test days.

\subsection{Hybrids and combination}
\label{sec:hybrid_design}

Two further experiments examine whether the networks learn anything that the pipeline does not. In the statistical-core LSTM (SC-LSTM), the pipeline forecast enters as a fixed offset,
\begin{equation}
\log\mu^{\mathrm{SC}}_{t+1,s}=\log\mu_{t+1,s}+f_s\!\left(\mathcal{X}_{t-M+1:t}\right),
\label{eq:sclstm}
\end{equation}
and a full LSTM is trained to produce the correction $f$ under the negative-binomial likelihood, with the dispersion fixed at the pipeline's volume-group values and a ridge penalty on $f$ weighted toward low-volume stations; its output layer is initialized at zero so that training starts exactly from the statistical forecast. If the network finds signal that the pipeline lacks, $f$ moves away from zero and the log score improves; otherwise early stopping halts training near the initial point. In the combination experiment, the pipeline and the full LSTM are averaged with equal weights \citep{bates1969combination}, and the station--hour correlation of their residuals is reported. A correlation near one with no gain from averaging would indicate that the two models have learned the same signal. A moderate correlation with a gain would indicate partial complementarity, which may arise from complementary information or from differences in bias; the bias ratio is reported to help distinguish between the two.

%=====================================================================
\section{Results}
\label{sec:experiments}
%=====================================================================

\subsection{Station-level accuracy}
\label{sec:station}

\begin{figure}[htbp]\centering
\includegraphics[width=\linewidth]{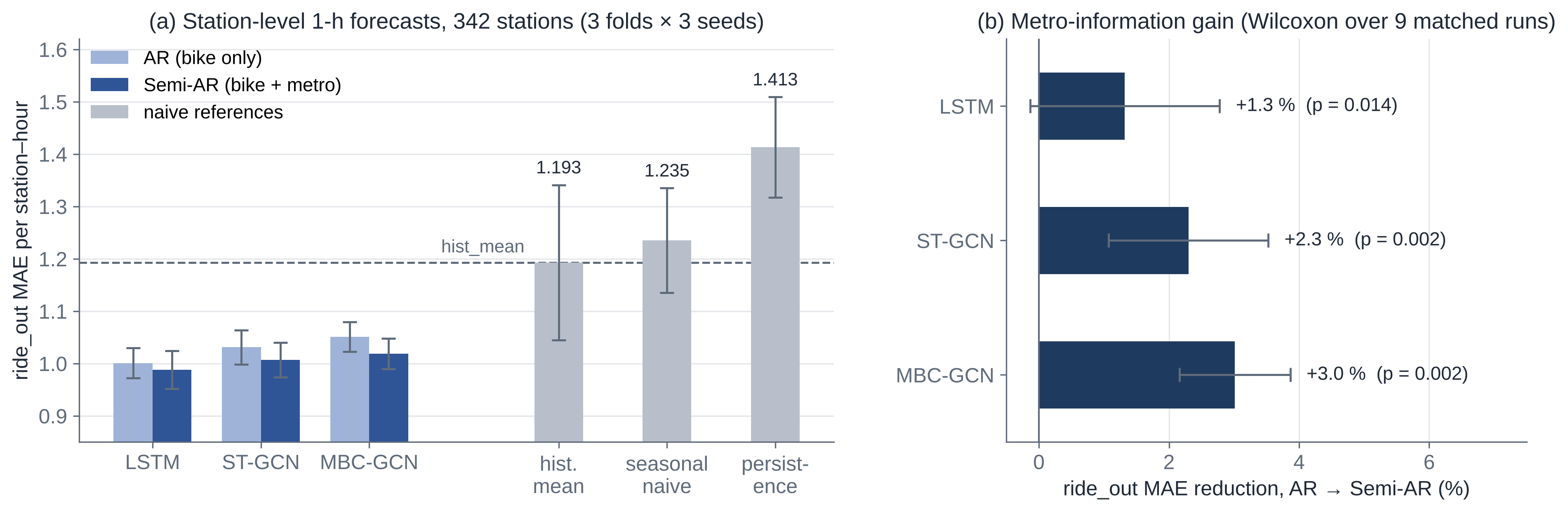}
\caption{Station-level results. (a) Departure MAE per station--hour for the three architectures under AR and Semi-AR inputs (dashed line: \texttt{hist\_mean}). (b) Reduction in departure MAE from adding metro flows (AR$\to$Semi-AR), with one-sided exact Wilcoxon $p$-values over nine matched runs.}
\label{fig:station}
\end{figure}

Table~\ref{tab:station} reports pooled station-level errors for the six deep configurations and the naive references, and Figure~\ref{fig:station}(a) shows the departure MAE.

\begin{table}[htbp]\centering\small
\begin{threeparttable}
\caption{Station-level one-hour-ahead forecasts, 342 stations, main panel; errors pooled over station--hours on held-out test days (3 folds $\times$ 3 seeds, mean $\pm$ s.d.). Semi-AR adds metro entry/exit flows to the inputs. Best value in each column in bold.}
\label{tab:station}
\begin{tabular}{llcccc}
\toprule
Model & Inputs & ride\_in MAE & ride\_out MAE & ride\_in $R^2$ & ride\_out $R^2$ \\
\midrule
LSTM    & Semi-AR & \textbf{0.978 $\pm$ 0.034} & \textbf{0.988 $\pm$ 0.036} & \textbf{0.647} & \textbf{0.607} \\
LSTM    & AR      & 0.992 $\pm$ 0.029 & 1.001 $\pm$ 0.029 & 0.629 & 0.586 \\
ST-GCN  & Semi-AR & 0.999 $\pm$ 0.036 & 1.007 $\pm$ 0.033 & 0.632 & 0.580 \\
ST-GCN  & AR      & 1.019 $\pm$ 0.037 & 1.031 $\pm$ 0.033 & 0.613 & 0.550 \\
MBC-GCN & Semi-AR & 1.016 $\pm$ 0.036 & 1.019 $\pm$ 0.029 & 0.617 & 0.569 \\
MBC-GCN & AR      & 1.047 $\pm$ 0.030 & 1.051 $\pm$ 0.028 & 0.581 & 0.520 \\
\midrule
\multicolumn{2}{l}{\texttt{hist\_mean}}      & 1.263 $\pm$ 0.184 & 1.193 $\pm$ 0.148 & 0.363 & 0.427 \\
\multicolumn{2}{l}{\texttt{seasonal\_naive}} & 1.253 $\pm$ 0.094 & 1.235 $\pm$ 0.100 & 0.383 & 0.339 \\
\multicolumn{2}{l}{\texttt{persistence}}     & 1.407 $\pm$ 0.091 & 1.413 $\pm$ 0.096 & 0.223 & 0.122 \\
\bottomrule
\end{tabular}
\end{threeparttable}
\end{table}

Every learning model improves on the best naive reference, by 12--17\% for departures and 16--22\% for arrivals. The differences among the learning models are much smaller. The best and worst of the six deep configurations differ by 0.06 trips per station--hour, and the three architectures with metro inputs (LSTM 0.988, ST-GCN 1.007, MBC-GCN 1.019) differ by 3\%, which is the size of the standard deviation across the nine runs; the ranking of the architectures also changes from fold to fold. The statistical pipeline matches the best of them: on the 309 non-empty stations on which it is evaluated (Table~\ref{tab:hybrid}) its departure MAE is 1.068, against 1.072 for the LSTM with metro inputs on the same stations, and it is the only model that provides calibrated intervals (negative-binomial log score 1.474, 90\% interval coverage 0.96).

Table~\ref{tab:groups} groups the errors by station volume. At low-volume stations (training mean below 0.5 departures per hour) all models are within 0.02 of one another and of the statistical pipeline. At high-volume stations (above 2 per hour) the models separate: the pipeline is ahead of the LSTM (1.493 against 1.525), and this is also where the combination of Section~\ref{sec:hybrid} gains most. The pipeline is therefore as accurate as any deep model at low-volume stations, and the differences between models are confined to the high-volume group.

\begin{table}[htbp]\centering\small
\caption{Departure MAE by station volume group (non-empty stations, 3-fold average). Group boundaries are 0.5 and 2 departures per hour in the training period.}
\label{tab:groups}
\begin{tabular}{lcccc}
\toprule
Model & All & Small ($<0.5$/h) & Mid (0.5--2/h) & Large ($\ge2$/h) \\
\midrule
Statistical pipeline      & 1.068 & 0.332 & 0.834 & 1.493 \\
LSTM (full, Semi-AR)      & 1.072 & 0.312 & 0.818 & 1.525 \\
Equal-weight average      & 1.036 & 0.322 & 0.809 & 1.449 \\
\bottomrule
\end{tabular}
\end{table}

\subsection{The value of metro information}
\label{sec:metro}

Table~\ref{tab:metro} isolates the contribution of metro flows by pairing AR and Semi-AR runs with identical seeds and folds. At station level, metro flows reduce departure MAE by 1.3\% for the LSTM, 2.3\% for ST-GCN and 3.0\% for MBC-GCN; the gains are small but consistent, significant at $p\le0.014$ for each architecture over the nine matched runs. For the network total the same comparison gives 10--34\%: SARIMAX improves from 79 to 52 trips per hour, the LSTM from 75 to 61, ST-GCN from 70 to 63 and MBC-GCN from 80 to 66, and gradient boosting and ridge regression by about 10\%.

\begin{table}[htbp]\centering\small
\caption{Marginal contribution of metro flows (AR$\to$Semi-AR), departure MAE. Station level: nine matched runs, main panel; $p$ from one-sided exact Wilcoxon signed-rank tests. Network total: three folds, reduction computed from the three-fold mean MAEs. The last row isolates the pipeline's recent-level step, which plays the same role as metro inputs at the network scale.}
\label{tab:metro}
\begin{tabular}{llcccc}
\toprule
Scale & Model & AR & Semi-AR & Reduction & $p$ \\
\midrule
Station   & LSTM            & 1.001 & 0.988 & 1.3\% $\pm$ 1.5 & 0.014 \\
Station   & ST-GCN          & 1.031 & 1.007 & 2.3\% $\pm$ 1.2 & 0.002 \\
Station   & MBC-GCN         & 1.051 & 1.019 & 3.0\% $\pm$ 0.9 & 0.002 \\
\midrule
Network   & SARIMAX         & 79.1  & 52.0  & 34\% & -- \\
Network   & LSTM            & 75.0  & 60.6  & 19\% & -- \\
Network   & ST-GCN          & 70.2  & 63.4  & 10\% & -- \\
Network   & MBC-GCN         & 80.1  & 65.6  & 18\% & -- \\
Network   & Gradient boosting & 85.7 & 77.2 & 10\% & -- \\
Network   & Ridge           & 105.6 & 96.0  & 9\%  & -- \\
Network   & Pipeline: profile $\to$ profile $\times$ recent level & 101.4 & 50.3 & 50\% & -- \\
\bottomrule
\end{tabular}
\end{table}

The contrast between the two scales is the first main finding. At a single station, the dominant structure in bike demand is the intraday profile, which can already be recovered from the station's own bike history; metro flows add a contemporaneous signal of the same shape and therefore contribute little new information. At the network level, every model recovers the profile, and the difficulty lies in the level: daily totals fell by two thirds during the study period, and the historical mean overshoots the test days by 30\% (bias ratio 1.30). Metro throughput is a real-time measurement of city-wide activity that these models do not extract from the bike history. This explains why it helps every architecture at the network level, while the statistical step that scales the profile by recent bike activity (stage (c)) halves the network-level error without metro inputs. The gain from metro inputs therefore depends on how much of the relevant variation the target series already contains, which differs between the two scales.

\subsection{Network-total forecasts}
\label{sec:total}

\begin{figure}[p]\centering
\includegraphics[width=0.76\linewidth]{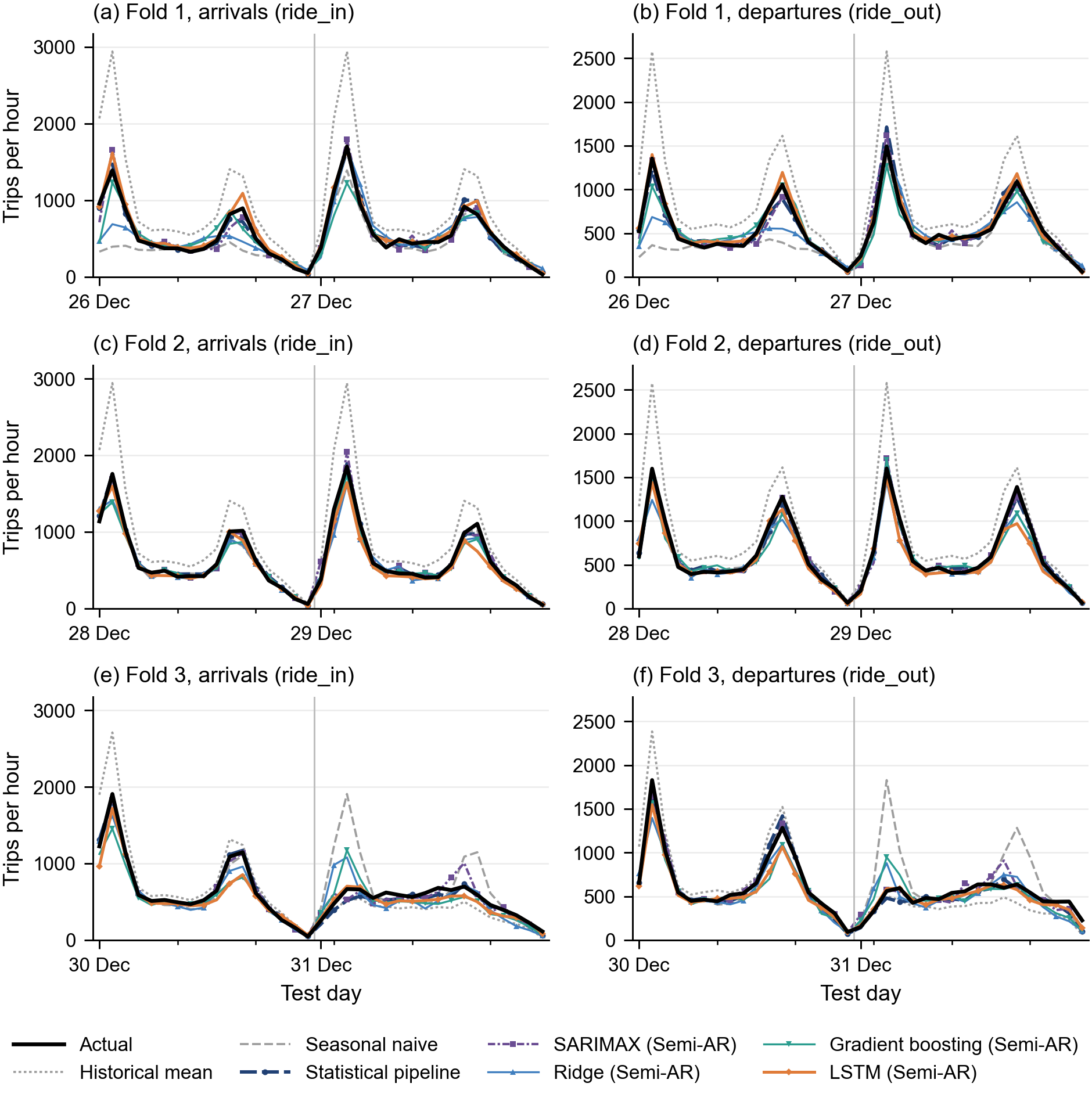}
\caption{Network-total forecasts: actual totals and the best model of each family over the three test windows, for arrivals (a, c, e) and departures (b, d, f); three-fold mean MAEs are given in Table~\ref{tab:total}.}
\label{fig:total}

\vspace{0.8em}
\includegraphics[width=0.76\linewidth]{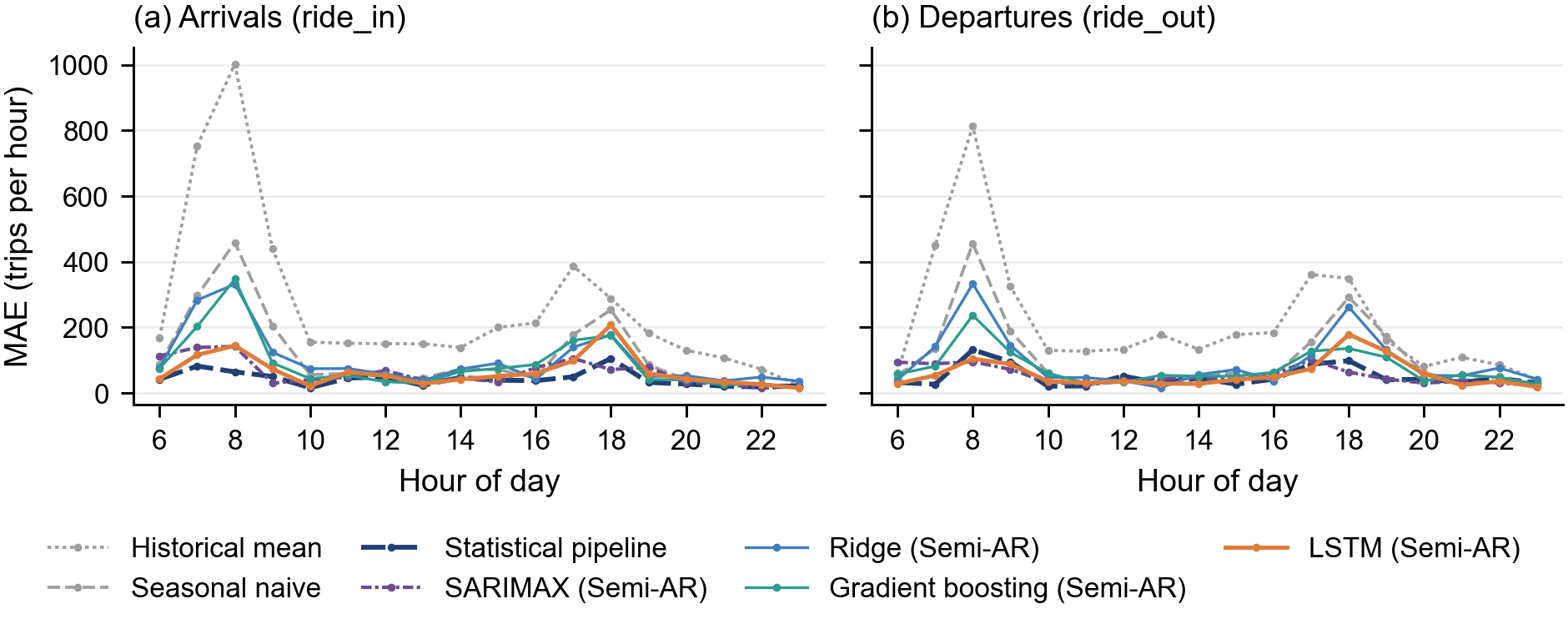}
\caption{Network-total MAE by hour of day, pooled over test days, for (a) arrivals and (b) departures; errors concentrate in the two commuting peaks.}
\label{fig:total_hourly}
\end{figure}

Table~\ref{tab:total} and Figure~\ref{fig:total} report one-hour-ahead forecasts of the network total (mean test-day demand 599 arrivals and 587 departures per hour). The ranking varies between folds (the LSTM and ST-GCN lead in fold 1, the pipeline in fold 2 and SARIMAX in fold 3), so the comparison rests on the three-fold averages. On that basis the statistical pipeline has the lowest mean error (departure MAE 50, MAPE 9.3\%, $R^2$ 0.96), followed closely by SARIMAX with metro inputs (52); the best deep model (LSTM, Semi-AR) is 20\% worse (61), and the tree and ridge models are 50--90\% worse. The three deep architectures are within one standard deviation of one another (61, 63 and 66, with fold-to-fold standard deviations of 13--15), and with only 105 test hours the gaps between the pipeline, SARIMAX and the LSTM are not statistically significant either (paired Wilcoxon over test hours, $p=0.90$ and $0.41$). With a single node the graph convolution reduces to a temporal convolution, so this is a comparison of temporal backbones on a series of 360 points, and on a series this short none of them has a lower mean error than the structured statistical model. Figure~\ref{fig:total_hourly} shows that the errors of all models concentrate in the morning (07--09h) and evening (17--19h) peaks, and that the statistical pipeline follows the peaks most closely.

\begin{table}[htbp]\centering\small
\caption{Network-total one-hour-ahead forecasts (three folds; deep models additionally three seeds). Reduction is relative to \texttt{hist\_mean}.}
\label{tab:total}
\resizebox{\linewidth}{!}{%
\begin{tabular}{llccccccc}
\toprule
Family & Model & ride\_in MAE & ride\_out MAE & ride\_out RMSE & MAPE (\%) & $R^2$ & Bias & vs.\ hist.\ mean \\
\midrule
Statistical & profile $\times$ recent level & \textbf{42.8} & \textbf{50.3 $\pm$ 8.3} & \textbf{66.7} & \textbf{9.3} & \textbf{0.96} & 0.98 & $-$77\% \\
SARIMA & SARIMAX, Semi-AR & 61.0 & 52.0 $\pm$ 10.2 & 72.9 & 10.5 & 0.95 & 0.99 & $-$76\% \\
Deep & LSTM, Semi-AR & 69.9 & 60.6 $\pm$ 15.4 & 87.2 & 10.0 & 0.93 & 0.95 & $-$73\% \\
Deep & ST-GCN, Semi-AR & 62.6 & 63.4 $\pm$ 12.9 & 89.1 & 10.9 & 0.93 & 0.93 & $-$71\% \\
Deep & MBC-GCN, Semi-AR & 68.8 & 65.6 $\pm$ 12.8 & 95.0 & 11.0 & 0.92 & 0.93 & $-$70\% \\
Tree & Gradient boosting, Semi-AR & 88.6 & 77.2 $\pm$ 9.4 & 107.9 & 13.4 & 0.89 & 0.95 & $-$65\% \\
Linear & Ridge, Semi-AR & 101.0 & 96.0 $\pm$ 14.4 & 143.7 & 17.0 & 0.81 & 0.92 & $-$57\% \\
Statistical & profile only & 104.3 & 101.4 $\pm$ 30.6 & 137.1 & 17.1 & 0.81 & 0.83 & $-$54\% \\
Naive & seasonal naive & 117.0 & 110.7 $\pm$ 47.8 & 199.4 & 19.2 & 0.52 & 0.94 & $-$50\% \\
Naive & historical mean & 264.3 & 220.9 $\pm$ 68.8 & 300.8 & 35.8 & 0.15 & 1.30 & 0 \\
Naive & persistence & 225.9 & 234.7 $\pm$ 39.1 & 333.0 & 45.9 & 0.06 & 1.00 & $+$6\% \\
\bottomrule
\end{tabular}}
\end{table}

\subsection{What the networks learn beyond the statistical forecast}
\label{sec:hybrid}

Table~\ref{tab:hybrid} reports the hybrid and combination experiments on the 309 non-empty stations (training mean of at least 0.05 departures per hour). The statistical-core LSTM barely changes its offset. The root-mean-square correction $f$ that it learns is 0.04 on the log scale (about $\pm4\%$), the validation log score moves from 1.3536 to 1.3530, and in six of nine runs the best epoch is one of the first three, so that training stops almost immediately. Its test MAE is 0.004 below that of the pipeline, a difference that is statistically detectable across stations but of no practical importance. The full LSTM trained from scratch reaches the same MAE as the pipeline (1.072 against 1.068); its forecasts correlate with those of the pipeline at 0.93, and the residuals of the two models correlate at 0.85. Averaging the two gives an MAE of 1.036, 3\% better than either model, with 87\% of stations improved ($p<10^{-30}$) and the largest gains at hubs (1.49$\to$1.45).

\begin{table}[htbp]\centering\small
\caption{Statistical core, LSTM, hybrid and combination on non-empty stations (3-fold average, departures). NLL = negative-binomial log score; cov.\ = 90\% interval coverage. Best value in each column in bold (bias ratio: closest to one).}
\label{tab:hybrid}
\begin{threeparttable}
\begin{tabular}{lccccccc}
\toprule
Model & MAE & RMSE & WMAPE & $R^2$ & Bias ratio & NLL & cov.\ 90\% \\
\midrule
Statistical pipeline              & 1.068 & 1.580 & 0.566 & 0.650 & \textbf{0.96} & 1.474 & 0.963 \\
LSTM (full, Semi-AR)              & 1.072 & 1.693 & 0.568 & 0.600 & 0.76 & -- & -- \\
SC-LSTM (offset + residual)       & 1.064 & 1.579 & 0.564 & 0.651 & 0.95 & 1.474 & 0.963 \\
Equal-weight average              & \textbf{1.036} & \textbf{1.566} & \textbf{0.549} & \textbf{0.657} & 0.86 & -- & -- \\
\bottomrule
\end{tabular}
\begin{tablenotes}\footnotesize\item Residual correlation (station--hour): pipeline vs.\ LSTM 0.853; pipeline vs.\ SC-LSTM 0.997. Forecast correlation pipeline vs.\ LSTM 0.931.\end{tablenotes}
\end{threeparttable}
\end{table}

\begin{figure}[htbp]\centering
\includegraphics[width=0.95\linewidth]{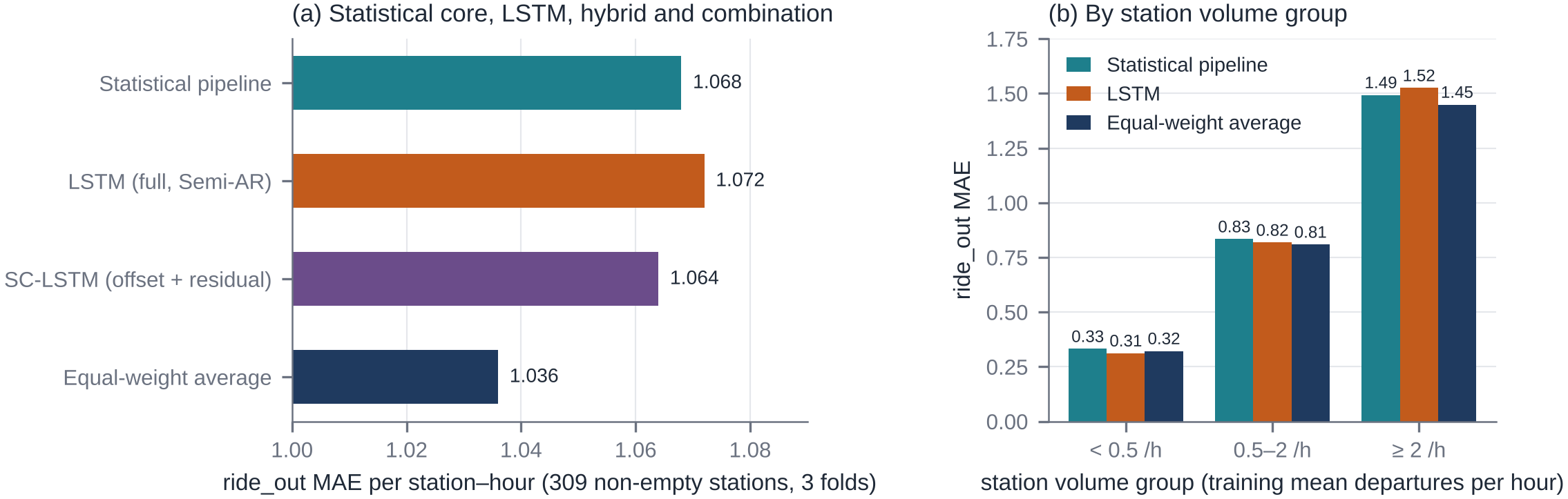}
\caption{Departure MAE of the statistical core, the LSTM, the hybrid and the combinations (left), and by station volume group (right).}
\label{fig:hybrid}
\end{figure}

These results indicate that, with the inputs and training used here, the network finds no predictable signal beyond the four-component pipeline, namely an intraday profile shared across stations and scaled by recent activity. The part of the residuals that is not shared (correlation 0.85) is what the average exploits, and Table~\ref{tab:hybrid} shows where it comes from. The LSTM forecasts sum to 76\% of actual demand, whereas the pipeline is nearly unbiased (0.96). A network trained with squared error on $\log(1+y)$ shrinks toward low counts. Because the absolute-error criterion rewards under-prediction of right-skewed counts (the MAE-optimal point forecast is the median, which lies below the mean), this bias is not visible in MAE, slightly improves it at low-volume stations (0.312 against 0.332), and appears only in RMSE, $R^2$, the bias ratio and the hour-of-day profile. Part of the combination gain is therefore a reduction of the network's downward bias rather than evidence of complementary information. For count data, MAE alone can thus rank a biased forecaster ahead of an unbiased one, and the bias ratio or a proper scoring rule should be reported with it.

\subsection{External comparison under a published protocol}
\label{sec:external}

Finally, we evaluate the statistical pipeline under the protocol of \citet{li2026cogclstm}, who used the same 18.06 million bike records and the same AFC period. That protocol attributes each record to the nearest station within 1~km (Euclidean), counts every record as a demand event, aggregates over the full day at 15, 30 and 60 minutes, and splits the 21 days into 17 for training and 4 for testing (29 December--1 January). We rebuilt the panel from the raw records under exactly these rules with 336 station locations, selected the pipeline's parameters on the last two training days and evaluated on the same four test days. The per-station mean counts of the rebuilt panel (7.3, 14.5 and 29.1 per step at 15, 30 and 60 minutes) match those implied by the published MAE and WMAPE (7.6, 15.1 and 30.2) to within 4\%, so the two panels are similar in overall count level. WMAPE is used as the primary metric because it is scale-free.

\begin{table}[htbp]\centering\small
\caption{Departure (unlock) forecasts under the 1-km protocol of \citet{li2026cogclstm}, test days 29 Dec--1 Jan, 336 stations. ``Poisson reference'' is the expected absolute error of a mean forecast under Poisson noise at the fitted rates of our panel. Published values are transcribed from Table~3 of that study (its 45-minute interval, not run here, is omitted); arrival (lock) values differ from the departure values by less than 0.003 in WMAPE and are shown in Figure~\ref{fig:li}(b). The three remaining published models (ARIMA, LightGBM, GRU) appear in Figure~\ref{fig:li} only.}
\label{tab:li}
\resizebox{\linewidth}{!}{%
\begin{tabular}{llccc|ccc|ccc}
\toprule
 & & \multicolumn{3}{c|}{15 min} & \multicolumn{3}{c|}{30 min} & \multicolumn{3}{c}{60 min} \\
Source & Model & MAE & RMSE & WMAPE & MAE & RMSE & WMAPE & MAE & RMSE & WMAPE \\
\midrule
This study & Poisson reference & -- & -- & 0.240 & -- & -- & 0.170 & -- & -- & 0.121 \\
This study & Historical average (HA) & 3.12 & 6.06 & 0.429 & 5.45 & 11.35 & 0.374 & 9.88 & 21.54 & 0.340 \\
This study & Seasonal naive & 2.96 & 5.19 & 0.408 & 4.68 & 8.70 & 0.322 & 7.58 & 15.21 & 0.261 \\
This study & Persistence & 2.75 & 4.51 & 0.379 & 4.57 & 7.83 & 0.314 & 10.07 & 18.85 & 0.346 \\
This study & Profile (a) & 2.33 & 3.90 & 0.320 & 3.69 & 6.41 & 0.254 & 6.44 & 11.61 & 0.221 \\
This study & Profile + EB (a+b) & 2.24 & 3.81 & 0.309 & 3.60 & 6.36 & 0.248 & 6.31 & 11.56 & 0.217 \\
This study & \textbf{Pipeline (a--c)} & \textbf{2.08} & \textbf{3.40} & \textbf{0.286} & \textbf{3.21} & \textbf{5.42} & \textbf{0.221} & \textbf{5.29} & \textbf{9.29} & \textbf{0.182} \\
\midrule
\citet{li2026cogclstm} & HA & 3.81 & 5.92 & 0.504 & 7.07 & 11.25 & 0.468 & 13.35 & 21.61 & 0.442 \\
\citet{li2026cogclstm} & XGBoost & 3.40 & 5.08 & 0.450 & 6.23 & 10.33 & 0.412 & 11.79 & 18.77 & 0.390 \\
\citet{li2026cogclstm} & LSTM & 3.73 & 5.09 & 0.494 & 5.99 & 10.57 & 0.392 & 13.39 & 19.68 & 0.443 \\
\citet{li2026cogclstm} & GCLSTM & 2.59 & 4.49 & 0.341 & 4.79 & 8.56 & 0.313 & 10.20 & 17.62 & 0.329 \\
\citet{li2026cogclstm} & CoGCLSTM & 2.60 & 4.45 & 0.342 & 4.66 & 8.22 & 0.305 & 9.53 & 16.81 & 0.308 \\
\bottomrule
\end{tabular}}
\end{table}

\begin{figure}[htbp]\centering
\includegraphics[width=\linewidth]{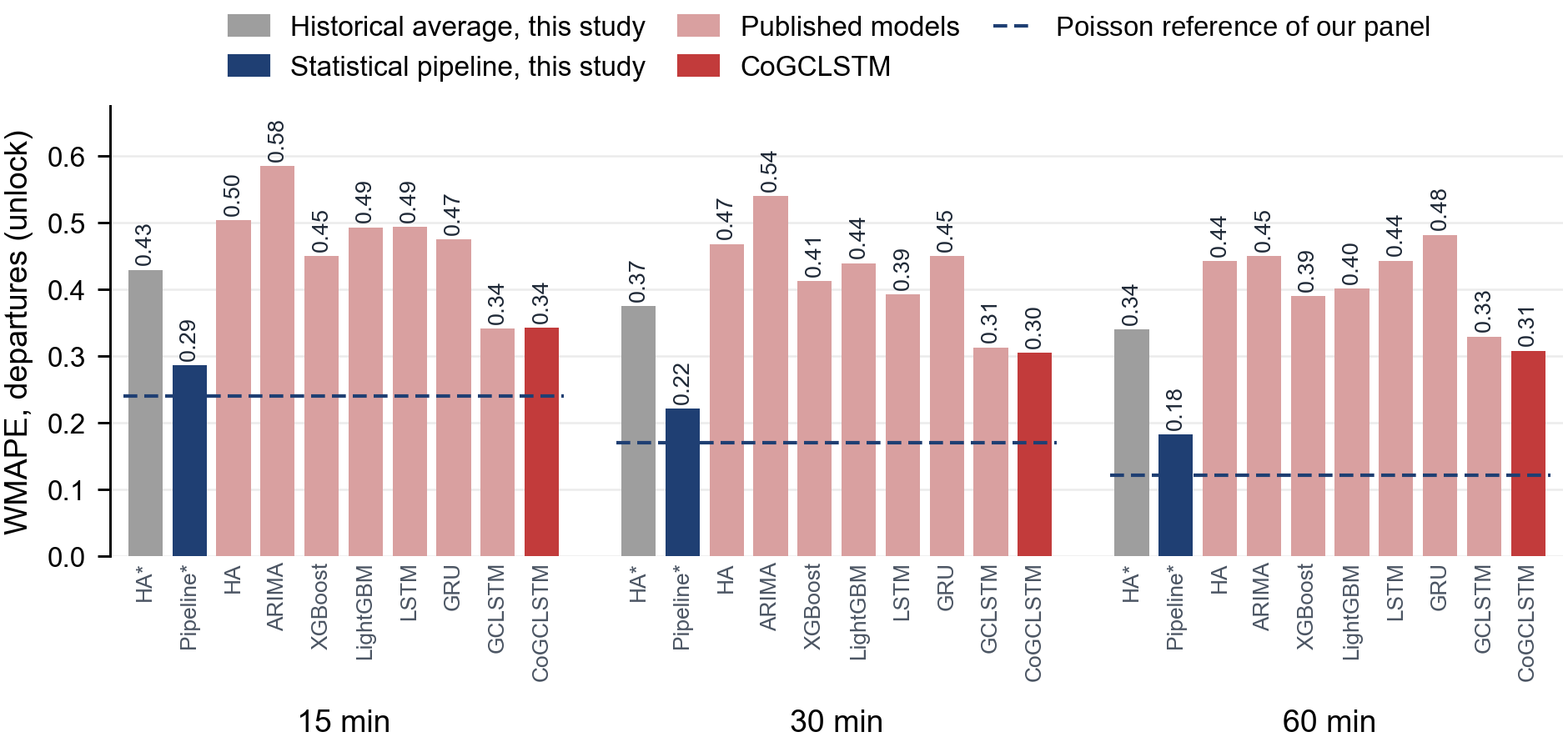}\\[2pt]
{\small (a) Departures (unlock)}\\[8pt]
\includegraphics[width=\linewidth]{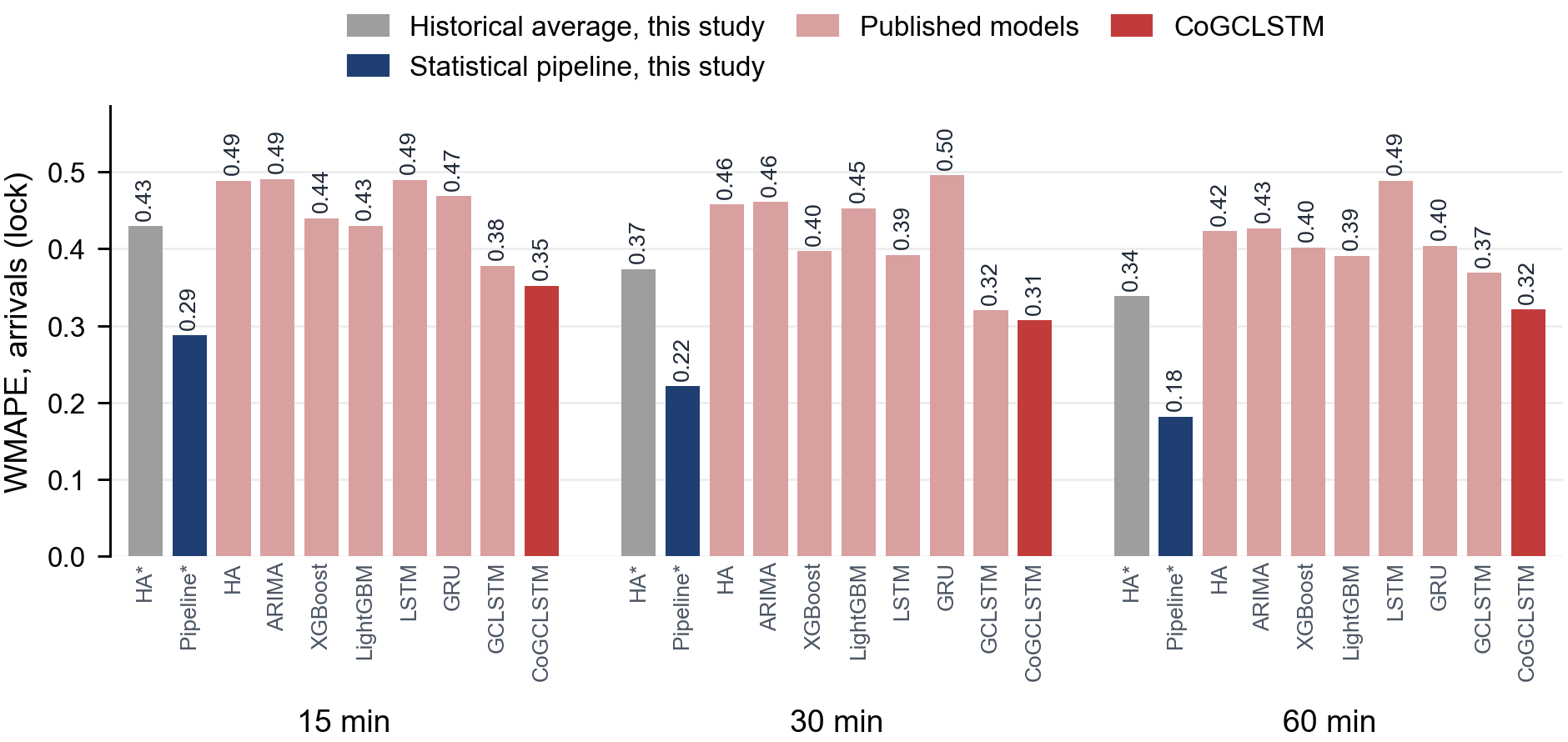}\\[2pt]
{\small (b) Arrivals (lock)}
\caption{WMAPE under the 1-km protocol for (a) departures and (b) arrivals: our historical average (gray) and statistical pipeline (blue) against the eight models of \citet{li2026cogclstm} (pink; red = CoGCLSTM). Dashed lines are the Poisson reference errors of our panel.}
\label{fig:li}
\end{figure}

Table~\ref{tab:li} and Figure~\ref{fig:li} give the comparison. The statistical pipeline attains departure WMAPEs of 0.286, 0.221 and 0.182 at 15, 30 and 60 minutes, against 0.342, 0.305 and 0.308 for the published CoGCLSTM, i.e.\ 16\%, 28\% and 41\% lower, and it is below each of the eight published models at every step. The arrival comparison (Figure~\ref{fig:li}b) is the same: the arrival WMAPEs of the pipeline (0.287, 0.221, 0.181) are within 0.003 of its departure values, whereas the published arrival errors are slightly higher than the departure ones (CoGCLSTM 0.35, 0.31, 0.32), so the margin is marginally wider for arrivals. MAE and RMSE show the same ordering, and the $R^2$ of the pipeline rises from 0.87 at 15 minutes to 0.93 at 60 minutes. Three points should be noted in interpreting this comparison. Our historical-average baseline (0.43, 0.37, 0.34) is stronger than the published one (0.50, 0.47, 0.44), and our seasonal-naive forecast at 60 minutes (0.261) is below the published CoGCLSTM (0.308), so that part of the advantage of the published model over its baselines reflects the weakness of those baselines rather than the architecture. The published errors also fall little as the aggregation window lengthens (0.34$\to$0.31$\to$0.31), whereas aggregation halves the Poisson reference (0.24$\to$0.12) and the error of the pipeline falls with it, remaining 0.05--0.06 above it at every step; an error that does not fall with aggregation suggests that systematic structure remains unexplained. Finally, the published model uses built-environment and weather inputs that are not used here, and 388 rather than 336 station locations. The published model was not re-run, so the gap is relative to its published scores and may partly reflect these differences.

%=====================================================================
\section{Discussion}
\label{sec:discussion}
%=====================================================================

The results give a conditional answer to the question in the title. Their implications for the use of deep learning in demand forecasting are discussed below, followed by the implications for policy and practice and the limitations of the study.

\subsection{Scale dependence of the value of metro information}
\label{sec:disc_scale}
Adding metro flows improved every model tested, whereas the choice between a graph and a recurrent backbone changed station-level error by at most 3\%, within run-to-run variability. The value of the metro signal is not fixed, however. It is 1--3\% where the target series already contains the shared intraday profile (a single station) and 10--34\% where the target depends on day-level activity that the tested models do not recover from bike history (the network total). This helps to reconcile the positive association between metro ridership and bike-sharing use reported in the literature \citep{ma2015bicycle,jin2025connecting} with the small station-level gain found here. The coupling is present at every scale, but its predictive value depends on how much of the relevant variation the target series already contains, and this differs between scales. For an operator, metro feeds therefore matter more for fleet-level than for station-level forecasts, and at the fleet level mainly for models that do not track the recent level explicitly.

\subsection{What the deep models learn}
\label{sec:disc_learn}
The hybrid experiment provides the most direct evidence on this question. Started from the statistical forecast, the network found almost nothing to add, and the residual correlation of 0.85 and the near-identical errors by station group point in the same direction. On this panel, the contribution of the graph convolution appears to be the pooling of stations toward a common shape, which empirical-Bayes shrinkage performs explicitly and in closed form, and the contribution of the recurrent layer appears to be the tracking of the recent level, which the two-window ratio performs with two parameters. A panel of 342 stations over 20 days with a two-thirds swing in demand seems to contain too little independent variation for the architectures to learn more than this structure. The gain from averaging the two families (3\%) is consistent and inexpensive, but part of it is the reduction of the downward bias of the network, which MAE conceals. For count data we therefore recommend reporting bias and a proper scoring rule alongside MAE, reporting a Poisson reference error so that the share of count noise in the error can be judged, and including a calibrated statistical pipeline among the baselines.

\subsection{Interpretation of published benchmark gains}
\label{sec:disc_bench}
Under the protocol of \citet{li2026cogclstm}, with per-station counts matching theirs to within 4\%, the pipeline is 16--41\% more accurate than the published graph model, and our historical average is 15--23\% more accurate than theirs. A 21-day sample is short, and the published model uses built-environment and weather information that is not used here, so the result does not indicate what graph-based models could achieve on longer samples. It does indicate that improvements attributed to architecture in this literature are, at least in part, improvements over baselines that can be substantially strengthened, and that the fixed-protocol, baseline-first evaluation that is now standard in general forecasting \citep{makridakis2020m4,hewamalage2023forecast} would also be useful in this field.

\subsection{Implications for policy and practice}
\label{sec:implications}

For operators, the results support a statistical pipeline as the default forecasting system for metro-adjacent bike demand. It runs in under a minute including tuning, requires no training infrastructure, yields calibrated intervals for stock-out risk, and is within 0.02 trips per station--hour of the best deep model at low- and mid-volume stations (Table~\ref{tab:groups}); every forecast decomposes into a usual profile, a shrinkage weight and a recent-level ratio that a dispatcher can inspect. Where accuracy at hubs justifies the effort, a jointly trained network should be combined with the pipeline rather than substituted for it. Whether a gain of 3\% at the busiest stations is worth a second system depends on how forecast error translates into rebalancing cost and lost trips, which differs between systems \citep{gammelli2022predictive,atacc2025bike} and was not modeled here. Operators should also monitor bias as well as MAE: the forecasts of the LSTM sum to 76\% of observed departures at the same MAE as the pipeline (Table~\ref{tab:hybrid}), so an operator that stocked stations according to that forecast would provide for about a quarter fewer departures than occur.

For transit agencies and cities, metro throughput is most useful as a real-time indicator of fleet-level demand, where its value is largest and where it protects against level shifts (holidays, weather, public-health disruptions) that history-based models cannot anticipate; for station-level forecasts a metro data feed adds little. The directional asymmetry in the coupling diagnostics (the entry$\to$arrival elasticity exceeds the exit$\to$departure elasticity at every distance inside the coupling zone) further suggests that metro entry flows are the more informative of the two feeds, and that they can serve as a timely indicator of bike accumulation and parking pressure at station entrances during inbound peaks. Because the coupling is flat out to a nominal 200\,m and has halved by 300--400\,m (Table~\ref{tab:rings}), this is also the distance over which metro-related parking demand should be planned for, for instance when geo-fenced parking \citep{zhao2021geo} or the parking rules applied to dockless fleets \citep{gu2019dockless,brousseau2025harmonizing} are dimensioned. Data exchange between agencies and operators can be framed as reciprocal, since bike flows also help to predict metro flows at 39--90\% of stations (Table~\ref{tab:granger}).

For those who procure, regulate or review forecasting tools, the external comparison shows that claimed gains in accuracy need a standard of evidence: a tuned statistical baseline that combines an intraday profile with the recent level of demand, test days that were not used for tuning, the same information given to every model, and the bias ratio and a noise reference reported alongside MAE. None of these requires access to the internals of a proprietary model.

\subsection{Limitations}
\label{sec:limitations}

The study covers one city during the COVID-19 wave of December 2022; the relative findings (information versus architecture, pipeline versus network) are expected to transfer more readily than the absolute errors. The implications drawn above rest on predictive accuracy: the study did not observe the costs of either kind of system or simulate rebalancing, and the coupling diagnostics are predictive associations, not causal effects of metro service on bike use. The main panel uses nominal Web-Mercator distances (about 170~m on the ground for the 200~m catchment), which affects the catchment labels but not the model comparisons. The distance-band diagnostics use 336 stations and, at hourly resolution, only station-hours with complete 5-minute coverage, so their confidence intervals are wide. The external comparison relies on numbers transcribed from a published table and on a rebuilt panel whose station set (336) is smaller than the published one (388); the per-station means agree to within 4\%, but the published model was not re-run. The hyperparameters of the deep models were selected on the Semi-AR configuration and reused for AR, which may favor Semi-AR, and several selected values lie on the boundary of the search space (Table~\ref{tab:deep_hyper}). Finally, built-environment, weather and event information are not used here, and any of them could change the balance between the model families.

%=====================================================================
\section{Conclusions}
\label{sec:conclusion}
%=====================================================================

This study examined when metro passenger flows improve short-term shared-bike demand forecasts and what deep spatiotemporal models learn that a calibrated statistical model does not. Using 20 days of synchronized AFC and dockless-bike records for 342 Shanghai stations under a common rolling-origin protocol, metro information was found to improve every model, by 1--3\% at station level and by 10--34\% for the network total; the difference reflects how much of the relevant variation the target series already contains at each scale. Architecture mattered much less. The three deep models differed by at most 3\% at station level, and the LSTM, the most accurate of them, did not improve on the statistical pipeline at high-volume stations. A four-component statistical pipeline (recency-weighted profile, empirical-Bayes shrinkage, recent-level ratio and negative-binomial errors) matched the deep models at station level, had the lowest mean error for the network total and provided calibrated intervals. A network given the pipeline forecast as an offset learned almost no residual signal, the residuals of the two approaches correlated at 0.85, and averaging them gave a further 3\%, partly through a reduction of the downward bias of the network. Under the 1-km protocol of a companion graph-based study on the same data, the weighted absolute percentage error of the pipeline was 16--41\% lower at every aggregation step.

For this problem, we found no evidence that the deep models capture structure beyond that of the statistical pipeline: a shared intraday profile scaled by recent activity, with metro throughput as an early indicator of level shifts. This structure can be stated explicitly, and doing so yields a baseline whose error is below the published scores of the graph model. For practice, operators can adopt a transparent statistical pipeline as the default and monitor bias as well as error, and transit agencies and cities can treat metro ridership as a real-time indicator of fleet-level demand and of bike accumulation at station entrances. Deep models add value in combination with the pipeline, most of all at hubs, and future work should examine whether richer inputs (built environment, weather, events), longer samples and explicitly multimodal targets extend their advantage. Benchmarks in this area should in any case be reported against a baseline of this kind, with held-out test days, paired tests, bias and a noise reference, so that improvements can be attributed to information, architecture or protocol.

%=====================================================================
\appendix
% Appendix numbering: sections A, B; tables and equations A.1, B.1, ...;
% section headings read "Appendix A", "Appendix B".
\counterwithin{table}{section}
\counterwithin{equation}{section}
\makeatletter
\renewcommand{\@seccntformat}[1]{\@nameuse{appsecprefix@#1}\csname the#1\endcsname\quad}
\def\appsecprefix@section{\appendixname~}
\makeatother

\section{Statistical pipeline: derivations, algorithm and Poisson reference}
\label{app:stat}
%=====================================================================

\subsection{Gamma--Poisson derivation of the shrinkage estimator}
\label{app:stat_eb}
Let $\theta_{h,s}$ be the unknown mean count at hour $h$ of station $s$ on the target day $d$, let the observations $y_{i,h,s}$, $i\in\mathcal{D}_d$, be Poisson with mean $\theta_{h,s}$ and enter the likelihood with the recency weights $w_{d,i}$ of Eq.~(\ref{eq:profile}), and let $\theta_{h,s}\sim\mathrm{Gamma}(\kappa_{\mathrm{eb}},\kappa_{\mathrm{eb}}/m_{d,h,s})$ (shape, rate), so that $\mathbb{E}[\theta]=m$ and $\mathrm{CV}[\theta]=\kappa_{\mathrm{eb}}^{-1/2}$. The weighted likelihood $\prod_i\mathrm{Poisson}(y_{i,h,s};\theta)^{w_{d,i}}\propto\theta^{\sum_iw_{d,i}y_{i,h,s}}e^{-\theta\sum_iw_{d,i}}$ is conjugate, so the posterior is $\mathrm{Gamma}(\kappa_{\mathrm{eb}}+W_dp_{d,h,s},\ \kappa_{\mathrm{eb}}/m_{d,h,s}+W_d)$, using $\sum_iw_{d,i}y_{i,h,s}=W_dp_{d,h,s}$. Its mean is Eq.~(\ref{eq:eb}). The weight $\omega=W_d/(W_d+\kappa_{\mathrm{eb}}/m)$ follows by writing the ratio as $(W_dp+(\kappa_{\mathrm{eb}}/m)\,m)/(W_d+\kappa_{\mathrm{eb}}/m)$. The posterior treats hours independently given the prior; the correlation between adjacent hours is carried by the prior mean through the common shape $g$, which is estimated from all stations and all earlier days and is therefore known with little error. The prior precision $\kappa_{\mathrm{eb}}$ could be estimated by the method of moments from the marginal variance of the profiles \citep{clayton1987empirical}; it is instead selected on validation days so that the pipeline is tuned in the same way as the deep models.

\subsection{Conjugate derivation of the recent-level ratio}
\label{app:stat_ratio}
Within the current day let $Y_{\tau,s}\sim\mathrm{Poisson}(r\,b_{\tau,s})$ for $\tau$ in a window $\mathcal{W}(t)$, with $b_{\tau,s}$ known at $t$, and let $r\sim\mathrm{Gamma}(\rho,\rho)$, centered on one with $\mathrm{CV}=\rho^{-1/2}$. The likelihood in $r$ is $\propto r^{\sum y}e^{-r\sum b}$, so the posterior is $\mathrm{Gamma}(\rho+\sum_{\tau}y_{\tau,s},\ \rho+\sum_{\tau}b_{\tau,s})$ with mean equal to the pseudo-count ratio $\hat r_{\mathcal{W}}$ of Eq.~(\ref{eq:ratio_window}), for $\mathcal{W}$ either $\mathcal{W}_{\mathrm{day}}$ or $\mathcal{W}_{3\mathrm{h}}$; the maximum-likelihood ratio $\sum y/\sum b$ is the $\rho\to0$ limit. Setting $\rho'=3\rho/H$ for the three-hour window makes the prior worth the same fraction of expected trips in both windows. The log-linear pool $\log r=(1-\beta)\log\hat r_{\mathrm{day}}+\beta\log\hat r_{3\mathrm{h}}$ is the geometric mean for $\beta=1/2$ and is symmetric under $r\mapsto1/r$, which an arithmetic pool is not.

\subsection{Negative-binomial distribution and the Poisson reference}
\label{app:stat_nb}
The NB2 probability mass function with mean $\mu$ and dispersion $\phi$ is
\begin{equation}
\Pr(Y=y)=\frac{\Gamma(y+\phi)}{\Gamma(\phi)\,y!}\Big(\frac{\phi}{\phi+\mu}\Big)^{\phi}\Big(\frac{\mu}{\phi+\mu}\Big)^{y},\qquad
\mathrm{Var}[Y]=\mu+\mu^2/\phi,
\label{eq:nbpmf}
\end{equation}
which reduces to the Poisson as $\phi\to\infty$ \citep{cameron2013regression,hilbe2011negative}. With $\mu$ fixed at the pipeline forecasts, the log-likelihood is maximized over $\log\phi$ by a bounded one-dimensional search per volume group. The logarithmic score $-\log\Pr(Y=y)$ is strictly proper \citep{gneiting2007strictly}. If a count were Poisson with known mean $\mu$, the expected absolute error of the mean forecast is
\begin{equation}
\mathbb{E}|Y-\mu|=\frac{2\,e^{-\mu}\,\mu^{\lfloor\mu\rfloor+1}}{\lfloor\mu\rfloor!}
\label{eq:floor}
\end{equation}
\citep{ramasubban1958mean}. Averaged over station--hours at the fitted rates, this is the Poisson reference of Section~\ref{sec:protocol}, against which model errors are reported; it is the error of the mean forecast under Poisson noise and not a lower bound on MAE.

\subsection{Algorithm}
\label{app:stat_alg}

\begin{algorithm}[H]
\caption{Statistical pipeline: one-step-ahead forecast for all stations at origin $t=(d,h)$.}
\label{alg:pipeline}
\begin{algorithmic}[1]
\Require counts $y_{\tau,s}$ for $\tau\le t$; hyperparameters $(\lambda,\delta,\kappa_{\mathrm{eb}},\rho,\beta)$; dispersion $\phi$ by volume group
\State $d'\gets$ day of $t+1$; $h'\gets$ hour of $t+1$
\State $\mathcal{D}_{d'}\gets\{i<d':\mathrm{type}(i)=\mathrm{type}(d')\}$;\quad $w_{d',i}\gets 2^{-(d'-i-1)/\lambda}$;\quad $W_{d'}\gets\sum_{i\in\mathcal{D}_{d'}}w_{d',i}$
\For{each station $s$ and hour $h''=1,\dots,H$} \Comment{stage (a): profile of day $d'$ from days $<d'$}
  \State $p_{d',h'',s}\gets\sum_{i\in\mathcal{D}_{d'}}w_{d',i}\,y_{i,h'',s}\,/\,W_{d'}$;\quad blend with the all-day profile using $\delta$ pseudo-days
\EndFor
\State $\ell_{d',s}\gets\frac1H\sum_{h''}p_{d',h'',s}$;\quad $g_{d',h''}\gets\sum_sp_{d',h'',s}\big/\frac1H\sum_{h'''}\sum_sp_{d',h''',s}$ \Comment{stage (b): prior mean, Eq.~(\ref{eq:prior})}
\For{each station $s$ and hour $h''$}
  \State $m\gets\ell_{d',s}\cdot g_{d',h''}$;\quad $b_{d',h'',s}\gets(\kappa_{\mathrm{eb}}+W_{d'}\,p_{d',h'',s})/(\kappa_{\mathrm{eb}}/m+W_{d'})$ \Comment{Eq.~(\ref{eq:eb})}
\EndFor
\For{each station $s$} \Comment{stage (c): level of the current day from $\tau\le t$}
  \State $\hat r_{\mathrm{day}}\gets\big(\sum_{\tau=t-H+1}^{t}y_{\tau,s}+\rho\big)/\big(\sum_{\tau=t-H+1}^{t}b_{\tau,s}+\rho\big)$ \Comment{day window}
  \State $\hat r_{3\mathrm{h}}\gets\big(\sum_{\tau=t-2}^{t}y_{\tau,s}+\rho'\big)/\big(\sum_{\tau=t-2}^{t}b_{\tau,s}+\rho'\big)$ \Comment{3-h window, $\rho'=3\rho/H$}
  \State $r_{t,s}\gets\hat r_{\mathrm{day}}^{\,1-\beta}\cdot\hat r_{3\mathrm{h}}^{\,\beta}$;\quad $\mu_{t+1,s}\gets b_{d',h',s}\cdot r_{t,s}$
\EndFor
\State \Return $\mu_{t+1,\cdot}$ and predictive distributions $\mathrm{NB}(\mu_{t+1,s},\phi_{\mathrm{group}(s)})$ \Comment{stage (d)}
\end{algorithmic}
\end{algorithm}

%=====================================================================
\section{Hyperparameter search spaces}
\label{app:hyper}
%=====================================================================

\subsection{Statistical pipeline}
\label{app:grid}
The five hyperparameters are selected by exhaustive search over the grid in Table~\ref{tab:grid_stat}, keeping the combination with the smallest Poisson deviance (\ref{eq:deviance}) on training and validation days; test days are never used. The network-total series has no shrinkage step (a single series has no cross-station prior) and its pseudo-count $\rho$ is scaled to the aggregate level (about $10^4$ trips per day). The dispersion $\phi$ is not searched: with $\mu$ fixed, it is estimated per volume group by maximum likelihood with a bounded search over $\log\phi\in[-3,7]$. Selected values are given in Table~\ref{tab:stat_hyper}.

\begin{table}[htbp]
\centering\small
\caption{Search grid of the statistical pipeline.}
\label{tab:grid_stat}
\begin{threeparttable}
\begin{tabular}{llll}
\toprule
Symbol & Stage & Station level (200-m panel) & Network total \\
\midrule
$\lambda$ (half-life, days)        & (a) & $\{1,2,3,5,\infty\}$      & $\{1,2,3,5,\infty\}$ \\
$\delta$ (cold-start pseudo-days)  & (a) & $\{0,2,8\}$               & $\{0,1,2,4,8\}$ \\
$\kappa_{\mathrm{eb}}$ (prior precision) & (b) & $\{1,3,10,30,100\}$ & -- \\
$\rho$ (day-window pseudo-count)   & (c) & $\{5,20,40,80\}$          & $\{0,500,2000,5000\}$ \\
$\beta$ (3-h window weight)        & (c) & $\{0,0.25,0.5,0.75,1\}$   & $\{0,0.25,0.5,0.75,1\}$ \\
$\phi$ (NB2 dispersion)            & (d) & \multicolumn{2}{l}{maximum likelihood, $\log\phi\in[-3,7]$} \\
\midrule
Combinations & & 1{,}500 & 500 \\
\bottomrule
\end{tabular}
\begin{tablenotes}\footnotesize
\item $\rho'=3\rho/H$ is derived from $\rho$. The 1-km panel (Section~\ref{sec:external}) uses $\lambda\in\{1,2,3,5,\infty\}$, $\delta\in\{0,2,8\}$, $\kappa_{\mathrm{eb}}\in\{1,3,10,30\}$, $\rho\in\{0,5,20,50\}$ and $\beta\in\{0,0.25,0.5,0.75\}$ (960 combinations), searched once per aggregation interval.
\end{tablenotes}
\end{threeparttable}
\end{table}

\subsection{Deep models}
\label{app:hyper_deep}
Hyperparameters were selected with Optuna (Tree-structured Parzen estimator) on validation days only, using the Semi-AR configuration of each architecture; the selected values were then used for both input configurations. Table~\ref{tab:deep_hyper} lists the search spaces and the selected values. At station level, 15 trials for the LSTM and 20 for each graph model were run on folds 1 and 3 with two seeds each, minimizing the validation mean squared error on the standardized $\log(1+y)$ scale averaged over the four runs. For the network total, 25 trials per model were run on all three folds with two seeds, minimizing the validation MAE on the count scale. Training follows Section~\ref{sec:training}, with gradient clipping at 5.

\begin{table}[htbp]\centering\small
\begin{threeparttable}
\caption{Search spaces and selected hyperparameters of the deep models.}
\label{tab:deep_hyper}
\begin{tabular}{llccc}
\toprule
Hyperparameter & Search space & LSTM & ST-GCN & MBC-GCN \\
\midrule
\multicolumn{5}{l}{\emph{Station level (hourly, 342 stations)}} \\
Hidden width     & $\{64,96,128,160\}$ & 64 & 160 & 160 \\
Layers           & $\{1,2,3\}$ & 1 & 3 & 2 \\
Dropout          & $\mathcal U(0,0.4)$ & 0.007 & 0.33 & 0.39 \\
Learning rate    & $\log\mathcal U(10^{-4},3\times10^{-3})$ & $9.8\times10^{-4}$ & $7.2\times10^{-4}$ & $2.8\times10^{-3}$ \\
Weight decay     & $\log\mathcal U(10^{-7},10^{-3})$ & $7.9\times10^{-7}$ & $5.3\times10^{-6}$ & $2.2\times10^{-6}$ \\
Batch size       & $\{16,32,64\}$ & 16 & 32 & 16 \\
Sequence length  & $\{4,8,12\}$ & 8 & 12 & 4 \\
Chebyshev order  & $\{2,3,4\}$ & -- & 4 & 3 \\
\midrule
\multicolumn{5}{l}{\emph{Network total (hourly, one series)}} \\
Hidden width     & $\{16,32,64,128\}$ & 128 & 128 & 64 \\
Layers           & $\{1,2,3\}$ & 1 & 3 & 3 \\
Dropout          & $\mathcal U(0,0.5)$ & 0.19 & 0.003 & 0.048 \\
Learning rate    & $\log\mathcal U(3\times10^{-4},10^{-2})$ & $5.4\times10^{-3}$ & $5.6\times10^{-3}$ & $8.4\times10^{-4}$ \\
Weight decay     & $\log\mathcal U(10^{-7},10^{-2})$ & $1.1\times10^{-3}$ & $4.8\times10^{-6}$ & $2.2\times10^{-3}$ \\
Batch size       & $\{8,16,32\}$ & 8 & 8 & 32 \\
Sequence length  & $\{4,8,12,18\}$ & 18 & 8 & 12 \\
\bottomrule
\end{tabular}
\begin{tablenotes}\footnotesize
\item Notes: $\mathcal U$ and $\log\mathcal U$ denote uniform and log-uniform sampling. The Chebyshev order applies to the graph models only and is fixed at 1 for the network total, where all orders coincide on a single node. The time-embedding size of MBC-GCN is fixed at 16 (station level) and 8 (network total).
\end{tablenotes}
\end{threeparttable}
\end{table}

For the network-total baselines, the ridge penalty was chosen from $\{0.01,0.1,1,10,100,1000\}$; gradient boosting was tuned with 40 trials over learning rate $\log\mathcal U(0.01,0.3)$, iterations $\{100,200,400,800\}$, leaf nodes $\{4,8,16,31\}$, minimum leaf size $\{5,10,20,40\}$ and $\ell_2$ regularization $\log\mathcal U(10^{-3},10)$; SARIMAX used order $(2,0,1)(1,1,0)_H$ on the $\log(1+y)$ scale. The statistical-core LSTM (Section~\ref{sec:hybrid_design}) was tuned with 15 trials over hidden width $\{32,64,128\}$, dropout $\mathcal U(0,0.4)$, learning rate $\log\mathcal U(3\times10^{-4},5\times10^{-3})$, weight decay $\log\mathcal U(10^{-7},10^{-3})$, correction penalty $\log\mathcal U(0.1,30)$, penalty exponent $\{0,0.5,1\}$ and sequence length $\{4,8,12\}$.

% \section*{Acknowledgments}
% This work is supported by the Geospatial Center and the President's Research Fund at Saint Louis University. The experiment code is available at \url{https://github.com/lilab2024/metro-bike-forecast}.

% \section*{Declaration of generative AI and AI-assisted technologies in the writing process}
% The authors used Anthropic Claude for language polishing and code review. After using these tools, the authors reviewed and edited the content as needed and take full responsibility for the content of the manuscript.

% \section*{Declaration of competing interest}
% The authors declare that they have no known competing financial interests or personal relationships that could have appeared to influence the work reported in this paper.

% arXiv does not run BibTeX: upload the .bbl produced by this file (e.g. main.bbl) together with main.tex
\bibliographystyle{elsarticle-harv}
\bibliography{ref}

\end{document}